\documentclass[12pt]{article}
\usepackage[english]{babel}
\usepackage{dsfont}
\usepackage[utf8]{inputenc}
\usepackage[letterpaper,top=2cm,bottom=2cm,left=3cm,right=3cm,marginparwidth=1.75cm]{geometry}
\usepackage{physics}
\usepackage{amsmath}
 \usepackage{textcomp}
\usepackage{amssymb}
\usepackage{mathtools}
\usepackage{graphicx,booktabs,array}
\usepackage[colorlinks=true, allcolors=blue]{hyperref}
\usepackage[utf8]{inputenc}
\usepackage[english]{babel}
\usepackage{csquotes}
\usepackage[backend=bibtex, style=nature]{biblatex}

\AtEveryBibitem{%
  \clearfield{urldate}%
  \clearfield{month}
  \clearfield{eventdate}
  \clearfield{origdate}
}

\usepackage{tabularx}
\usepackage{geometry}  
\usepackage{lscape}    
\usepackage{amsmath}   
\usepackage{caption}
\usepackage{subcaption}
\usepackage[table]{xcolor}
\usepackage[normalem]{ulem}
\usepackage{booktabs}
\usepackage{tikz}
\usepackage{adjustbox}
\usepackage{rotating}
\usepackage{algorithm}
\usepackage{algpseudocode}
\usepackage[T1]{fontenc}
\usepackage{adjustbox}  
\usepackage{authblk}
\usepackage{array}
\usepackage{makecell}
\usepackage[utf8]{inputenc}
\usepackage{ulem} 
\usepackage[colorlinks=true, allcolors=blue]{hyperref}
\usepackage{soul}
\usepackage{hyperref}
\usepackage{pifont}
\newcommand{\cmark}{\ding{51}}%
\newcommand{\xmark}{\ding{55}}%

\title{MoCA: Multi-modal Cross-masked Autoencoder for Time Series in Digital Health}

\author[1]{Howon Ryu}
\author[2]{Yuliang Chen}
\author[3]{Yacun Wang}
\author[1]{Andrea Z. LaCroix}
\author[4]{Chongzhi Di}
\author[1]{Loki Natarajan}
\author[5]{Yu Wang}
\author[1]{Jingjing Zou}
\affil[1]{Herbert Wertheim School of Public Health and Human Longevity Science, University of California, San Diego, San Diego, CA, USA}
\affil[2]{Hal{\i}c{\i}o\u{g}lu Data Science Institute, University of California San Diego, San Diego, CA, USA}
\affil[3]{Department of Computer Science and Engineering, University of California San Diego, San Diego, CA, USA}
\affil[4]{Division of Public Health Sciences,
Fred Hutchinson Cancer Center, Seattle, WA, USA}
\affil[5]{Independent Researcher, Mountain View, CA, USA}

\newcommand{\edit}[1]{\textit{\color{green}}}
\usepackage{amsthm}

\newtheorem{assumption}{Assumption}

\theoremstyle{remark}
\newtheorem{remark}{Remark}

\begin{document}

\date{} 
\maketitle
\begingroup
\renewcommand\thefootnote{*}

\renewcommand\thefootnote{\textsuperscript{\dag}}

\endgroup

\begin{abstract}

Wearable devices enable continuous multi-modal physiological and behavioral monitoring, yet analysis of these data streams faces fundamental challenges including the lack of gold-standard labels and incomplete sensor data. While self-supervised learning approaches have shown promise for addressing these issues, existing multi-modal extensions present opportunities to better leverage the rich temporal and cross-modal correlations inherent in simultaneously recorded wearable sensor data. We propose the Multi-modal Cross-masked Autoencoder (MoCA), a self-supervised learning framework that combines transformer architecture with masked autoencoder (MAE) methodology, using a principled cross-modality masking scheme that explicitly leverages correlation structures between sensor modalities. MoCA demonstrates strong performance boosts across reconstruction and downstream classification tasks on diverse benchmark datasets. We further establish theoretical guarantees by establishing a fundamental connection between multi-modal MAE loss and kernelized canonical correlation analysis through a Reproducing Kernel Hilbert Space framework, providing principled guidance for correlation-aware masking strategy design. Our approach offers a novel solution for leveraging unlabeled multi-modal wearable data while handling missing modalities, with broad applications across digital health domains.

\end{abstract}

\section{Introduction}

Wearable and mobile device technology has transformed population health monitoring by enabling continuous, multi-modal measurement of physiological and behavioral signals in real-world environments. 
Modern wearable devices capture rich data streams from motion sensors such as accelerometers and gyroscopes \cite{GreenwoodHickman2021CHAP, stamatakis2022association, su2022epidemiology}, as well as emerging modalities such as electroencephalography (EEG) \cite{xie2022transformer, yuan2024eeg}, electrocardiography (ECG) \cite{yan2019fusing, hu2022transformer, meng2022enhancing, che2021constrained}, continuous glucose monitoring \cite{lee2023glucose, sergazinov2023gluformer}, and numerous other physiological sensors. 
Wearable devices have become instrumental in disease prevention \cite{javaid2022medicine, ginsburg2024key, menassa2025future}, early diagnosis and monitoring \cite{shajari2023emergence, etli2024future, mirjalali2022wearable}, and evidence-based clinical interventions \cite{gill2023potential, smuck2021emerging}, offering unprecedented opportunities to advance population health research and enable personalized medicine.

The field is moving toward comprehensive multi-modal monitoring via (i) identical sensors at multiple body sites (e.g., wrist/waist/hip), (ii) heterogeneous sensors integrated within a single device (e.g., accelerometer plus heart rate), and (iii) multiple devices used concurrently or sequentially, capturing interactions among activity, sedentary behavior, sleep, and physiological functions that single-sensor setups often miss.

However, progress in multi-modal wearable data analysis has been constrained by fundamental limitations that contrast with advances in other domains. Unlike computer vision and natural language processing, which benefit from large-scale labeled datasets, wearable health data analysis often relies on small clinical studies with missing gold-standard labels due to the challenges of laboratory calibration in free-living environments \cite{di2022considerations}. Additionally, missing data from one or more sensor modalities poses a common challenge in real-world scenarios, where device malfunctions, battery depletion, or user compliance issues can result in incomplete data streams.

Self-supervised learning (SSL) has emerged as a powerful solution to these data scarcity challenges by learning meaningful representations from unlabeled data \cite{xie2021self, li2021mst, caron2021emerging}, with success in time series \cite{time-transformer, data2vec, ts-contrastive, patchtst, xie2022simmim} and digital health applications \cite{krishnan2022self}. Masked autoencoders (MAE) \cite{mae} have proven especially effective within this paradigm by learning to reconstruct randomly masked portions of input data, successfully adapting to time series applications \cite{li2023ti, zhang2023trid, dong2023simmtm} with growing theoretical understanding \cite{cao2022understand, pan2022towards, zhang2022mask}. 
Critically, MAE's masking and reconstruction mechanism provides a natural solution to the missing data challenge by enabling models to impute missing sensor streams using information from available modalities, offering significant advantages over other SSL approaches that require complete data. However, extending MAE to multi-modal wearable data remains challenging due to the complex temporal correlations and cross-modal dependencies inherent in simultaneously recorded physiological signals.

In digital health applications, these correlations are often temporal in nature, with simultaneously measured signals across different modalities providing real-time insights into each other. For example, heart rate variability may correlate with movement patterns, while accelerometer data can provide context for interpreting physiological signals. While current multi-modal MAE approaches \cite{yang2024cmvim, bachmann2022multimae, yan2022multi} have made progress, there remain significant opportunities to better leverage these rich inter- and intra-modality correlation structures through principled masking strategies that are specifically designed to account for the interdependent nature of wearable device data.

We propose the \textbf{M}ulti-m\textbf{o}dal \textbf{C}ross-masked \textbf{A}utoencoder (MoCA), a novel approach that incorporates both intra- and inter-modality correlation structures in its masking strategy. Our contributions can be summarized into the following:
\begin{itemize}
    
\item We present MoCA, a self-supervised learning (SSL) framework that combines the state-of-the-art transformer architecture with masked autoencoder (MAE) methodology for multi-modal time series pre-training, employing a principled cross-modality masking strategy that randomly masks across time windows and sensor axes of all modalities to encourage the model to learn inter- and intra-modality correlation structures.

\item We empirically demonstrate that MoCA consistently outperforms existing baselines through extensive experiments on real-world physical activity data, transfer learning across multiple external datasets, comprehensive ablation studies, and interpretable visualizations. MoCA demonstrates strong performance in downstream classification tasks. On UCI-HAR \cite{ucihar}, it achieves 96.60\% accuracy with random sample splits (subjects may appear in both training and test sets) and 91.36\% under subject-wise cross-validation (training and test sets contain distinct subjects), improving upon the supervised ViT-Base model by 5.2 percentage points.  

\item MoCA achieves state-of-the-art transfer learning across six external datasets. In both LP and FT settings, MoCA delivers the highest overall accuracies among all approaches compared. It reaches the highest linear probing accuracy on five datasets, raising the average to 69.8\% (+14.9 points over Yuan et al. \cite{harnet}); under fine-tuning, it leads on three datasets and attains the best average accuracy of 76.8\%, slightly surpassing Yuan et al. (76.5\%).

    \item We establish one of the first theoretical guarantees for multi-modal Masked Autoencoder (MAE) performance by demonstrating a fundamental connection between multi-modal MAE loss and kernel canonical correlation analysis (KCCA) through a Reproducing Kernel Hilbert Space (RKHS) framework, providing principled guidance for how correlation structures in multi-modal wearable data can guide optimal masking strategy design. The results set a foundation for future studies on development of efficient masking schemes adapted to different types of multi-modal wearable device data.
    
\end{itemize}

\section{Results}
\label{sec:4}



\begin{figure}[htbp]
    \centering
    \includegraphics[width=1\linewidth]{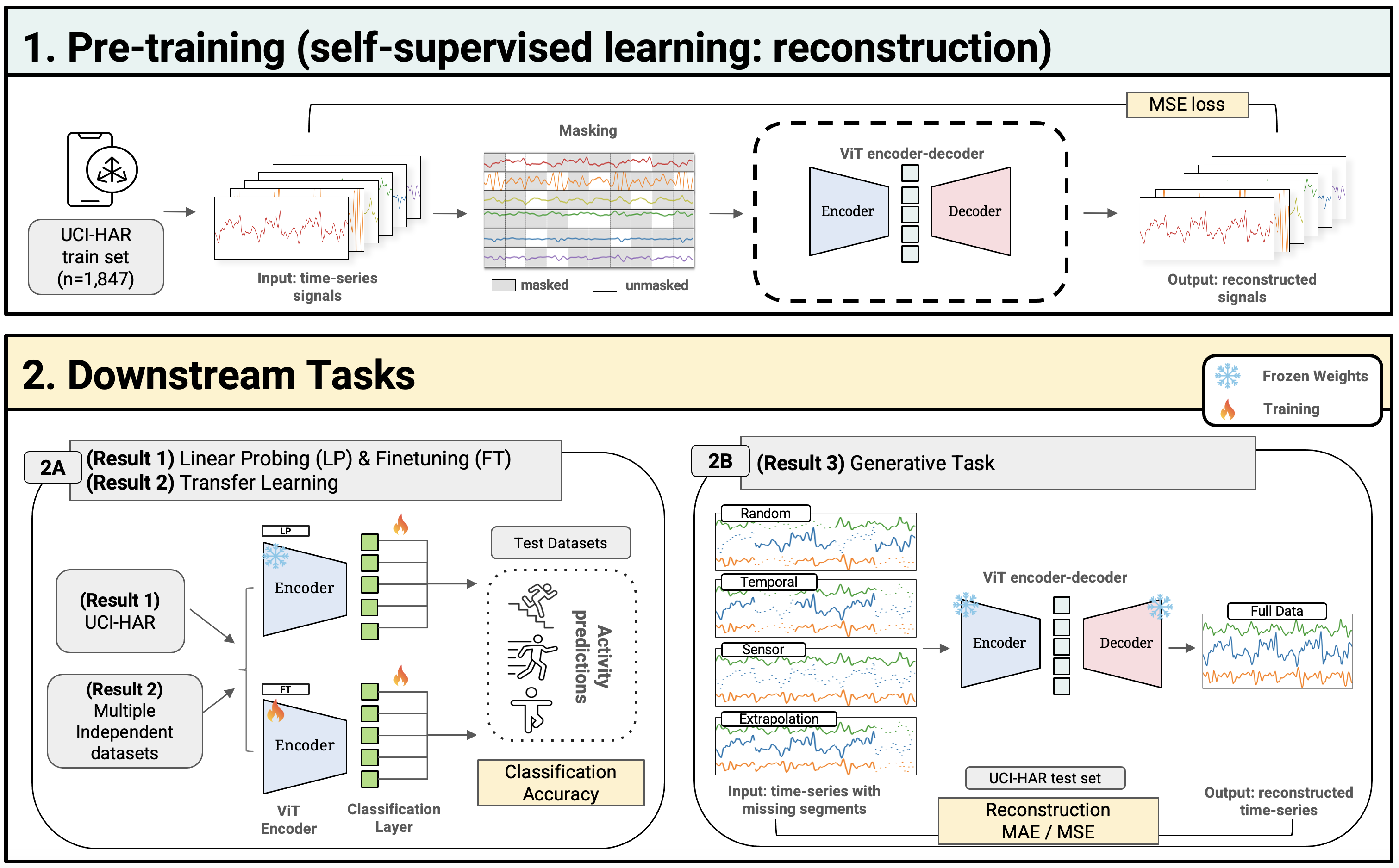}
    \caption{\textbf{MoCA pre-training and evaluation framework.} \textbf{[1. Pre-training]} MoCA is pre-trained on the UCI-HAR dataset \cite{ucihar} with and without data augmentation. Tri-axial accelerometer and gyroscope signals are used as input, with each axis treated as a distinct modality. The input signals are patchified and represented as tokens, followed by cross-modality masking with distinct tokens assigned to each modality. The patches go through the ViT-Base Autoencoder (ViT encoder-decoder) that reconstructs the complete original time series. Pre-training is carried out by optimizing the reconstruction loss, calculated as the mean squared error (MSE) between the original and reconstructed signals on both masked and unmasked patches. The resulting weights obtained during pre-training are transferred to downstream tasks.
    \textbf{[2A. Downstream Tasks: Activity Classification]} The encoder weights learned from pre-training, with a newly attached classification head (discarding the pre-training decoder), are evaluated on the UCI-HAR dataset (Result 1 - LP \& FT) and on additional six unseen benchmark datasets (Result 2 - Transfer Learning). In linear probing (LP), learned encoder weights are frozen and only the classification head is trained on the training set. In fine-tuning (FT), both encoder and classification layer weights are updated during training. For both LP and FT, classification accuracy is measured on the held-out test set of each dataset. \textbf{[2B. Downstream Tasks: Reconstruction/Imputation of Missing Data]} In a separate downstream task, we assess models' performance of imputing missing data under four types of artificially generated missing data patterns. In the Random Imputation task ("Random" in figure), patches are removed at random positions across time and modalities. In the Temporal Imputation task ("Temporal"), the same random time intervals are removed across all modalities. In the Sensor Imputation task ("Sensor"), all time intervals of a randomly selected modality are removed. In the Temporal Extrapolation task ("Extrapolation"), the final time intervals across all modalities are removed to evaluate forecasting ability. Masked inputs are processed by the pre-trained ViT encoder–decoder to reconstruct (impute) the missing segments. Imputation accuracy is quantified using mean absolute error (MAE) and mean squared error (MSE) between the original and reconstructed signals.}
    \label{fig:fig1}
\end{figure}

The overall experimental framework of MoCA is demonstrated in Figure \ref{fig:fig1}. MoCA is pre-trained on the UCI-HAR dataset and evaluated through two downstream tasks:
\begin{enumerate}
    \item Classification task, where encoder weights from pre-training are either frozen for linear probing or used as initialization for fine-tuning with a classification head on UCI-HAR. Performance is further evaluated in a similar way through transfer learning on six benchmark datasets to measure cross-dataset generalization, with comparisons against both supervised and self-supervised baselines (transfer learning);
    \item Generative Task, where the pre-trained encoder–decoder reconstructs artificially masked inputs under four missing data scenarios. For this task, MoCA is compared with statistical methods, as existing supervised and self-supervised approaches do not support imputation.
\end{enumerate}




\subsection{Pre-training Models}
\label{pre-training}

Our proposed method, \textbf{MoCA}, is pre-trained using a ViT-Base encoder–decoder backbone (details in Section~\ref{model_arch}) on the UCI-HAR dataset with 4-second input windows. Training is performed for 4{,}000 epochs with 75\% mask ratio using the AdamW optimizer (initial learning rate $5\times10^{-4}$, batch size 50; additional details in Appendix~\ref{implementation}). To increase the diversity of the training sample, we apply augmentation strategies (Section~\ref{data_aug_sec}), expanding the data from 1{,}847 to 369{,}400 samples; when using the augmented set, training is limited to 20 epochs so that the total number of optimization steps matches the unaugmented setting. We evaluate MoCA with and without augmentation respectively.  

As a critical baseline, we implement \textbf{MAE Synchronized Masking}, a masked autoencoder framework~\cite{mae} that adopts a synchronized masking scheme: within each window, all modalities are masked or unmasked simultaneously. MoCA differs in its \emph{cross-modality masking} scheme, which allows different time patches to be masked across modalities, thereby promoting the learning of cross-modal dependencies.

The second baseline is \textbf{Yuan et al. 2024}~\cite{harnet}, a multi-task self-supervised framework using a 1D ResNet-18 feature extractor with built-in augmentations such as axis swaps and rotations, trained on large-scale UK Biobank accelerometer data ($>$700,000 person-days).
To ensure a fair comparison, we re-trained Yuan et al. on the accelerometer subset of the UCI-HAR training set using the same pre-training configuration as MoCA (w/o aug.): 50 Hz sampling rate, 4-second windows (zero-padded to 5-second windows), and identical training epochs. We did not apply our proposed augmentation, since Yuan et al.’s self-supervised objective already incorporates built-in augmentation; adding additional augmentation would alter the input distribution and compromise comparability. Our setup was therefore aligned as closely as possible with the original Yuan et al. framework.

\begin{table}[t]
\centering
\caption{\textbf{Activity classification on UCI-HAR data.}
Top-1 classification accuracy (\%) under fine-tuning (FT) and linear probing (LP): (a) Results on the standard sample-wise training/test split, and (b) Results from subject-wise five-fold cross-validation (mean $\pm$ standard deviation from the five folds). Data augmentation (aug.) is applied where indicated. LP is not applicable (/) for ViT-Base, since it is a supervised learning model where the encoder weights are trained from random initialization along with classification layer.}
\label{tab:main}
\begin{subtable}{0.75\columnwidth}
\centering
\caption{Classification with standard sample-wise training/test split}
\resizebox{\textwidth}{!}{
\begin{tabular}{l|c|cc}
\toprule
\textbf{Models} & \textbf{Pre-training} & \textbf{FT Top-1 (\%)} & \textbf{LP Top-1 (\%)} \\
\midrule
ViT-Base \cite{vit}& \xmark & 91.40 & / \\
MAE Synchronized Masking (w/o aug.)& \cmark & 91.70 & 86.30 \\
MAE Synchronized Masking (w/ aug.)& \cmark & 91.30 & 88.40 \\
\midrule
MoCA (w/o aug., Ours)& \cmark & 94.80 & 92.40 \\
\textbf{MoCA (w/ aug., Ours)}& \cmark & \cellcolor{gray!30}\textbf{96.60} & \cellcolor{gray!30}\textbf{93.10} \\
\bottomrule
\end{tabular}}
\end{subtable}

\vspace{1em}

\begin{subtable}{0.75\columnwidth}
\centering
\caption{Classification with subject-wise 5-fold cross-validation}
\resizebox{\textwidth}{!}{
\begin{tabular}{l|c|cc}
\toprule
\textbf{Models} & \textbf{Pre-training}& \textbf{FT Top-1 (\%)} & \textbf{LP Top-1 (\%)} \\
\midrule
ViT-Base \cite{vit} & \xmark & 86.20 $\pm$ 2.53 & / \\
MAE Synchronized Masking (w/o aug.) & \cmark & 91.08 $\pm$ 1.56 & 86.16 $\pm$ 0.95 \\
\textbf{MoCA (w/o aug., Ours)} & \cmark & \cellcolor{gray!30}\textbf{91.36 $\pm$ 1.56} & \cellcolor{gray!30}\textbf{87.78 $\pm$ 2.87} \\
\bottomrule
\end{tabular}}
\end{subtable}

\end{table}

\subsection{Activity Classification on UCI-HAR}
\label{discriminative_task}

We first evaluate MoCA's downstream classification performance and compare with baseline models on the UCI-HAR dataset under two protocols: (a) the standard sample-wise training/test split, and (b) subject-wise five-fold cross-validation, which tests generalization to unseen participants. 
For each protocol, we report top-1 accuracy, which is the proportion of samples where the true label matches the model’s highest-probability prediction.
Each model is assessed under two training strategies: fine-tuning (FT), where both the encoder and classification head are updated during downstream training, and linear probing (LP), where the encoder weights learned from pre-training are kept frozen and only the classification head is updated during downstream training.


We compare MoCA's downstream FT/LP classification performances with the pre-trained model: MAE Synchronized Masking with and without data augmentation. In addition, we include a supervised ViT-Base \cite{vit} model (ViT-Base in Table~\ref{tab:main}) trained from random initialization (without pre-training) as a baseline. All classification models are further optimized for FT/LP tasks for 50 epochs; implementation details are provided in Appendix~\ref{implementation}.

Table~\ref{tab:main}(a) presents results on the standard sample-wise split, where training and test data are randomly split; subjects may appear in both training and test sets. MoCA substantially outperforms both baselines, achieving 96.6\% accuracy in FT and 93.1\% in LP with augmentation. Even without augmentation, MoCA exceeds MAE Synchronized Masking by more than three percentage points in FT and over four points in linear probing. These gains highlight the effectiveness of MoCA’s cross-modality masking in capturing dependencies across sensor streams. 
We note that augmentation further boosts performance, yielding the best results overall. 
Notably, MoCA also surpasses the fully supervised ViT-Base: its linear probing accuracy without augmentation already exceeds the supervised baseline, and with augmentation, MoCA FT reaches 96.6\% compared to 91.4\% for ViT-Base. This demonstrates that MoCA pre-training produces more transferable and generalizable representations than supervised learning from random initialization. Additional visualizations for the classification results including the confusion matrices can be found in Appendix \ref{supp_vis}.

Table~\ref{tab:main}(b) reports results from subject-wise five-fold cross-validation, where models are trained and tested on disjoint sets of participants (subject IDs).
This evaluation is conducted without data augmentation to single out the effect of subject-wise split. This protocol eliminates correlations between training and test samples of the same subjects and directly measures robustness to subject heterogeneity. Average top-1 accuracy across the five folds, along with standard deviation (mean $\pm$ standard deviation calculated from the five folds), is shown in Table~\ref{tab:main}(b). 

As expected, accuracies are lower than under the sample-wise split, reflecting the greater challenge of generalizing to unseen participants. Nevertheless, MoCA achieves the highest performance in both FT (91.36\%) and LP (87.78\%), outperforming MAE Synchronized Masking and the supervised ViT-Base baseline.

\subsection{Transfer Learning}
\label{transfer_learning}

Wearable devices used in clinical settings often vary in sensor modality, placement, and activity distribution. Motivated by this variability, we evaluate MoCA’s ability to generalize under challenging out-of-distribution (OOD) conditions.
Specifically, we evaluate transfer learning performance on six additional benchmark datasets through classification tasks: WISDM \cite{wisdm}, IMWSHA \cite{imwsha1,imwsha2}, OPPORTUNITY \cite{oppo}, ADL \cite{adl}, PAMAP2 \cite{pamap}, and RealWorld \cite{realworld}. MoCA was pre-trained on accelerometer and gyroscope data. Among the transfer datasets, only IMWSHA includes both modalities, whereas WISDM, OPPORTUNITY, ADL, PAMAP2, and RealWorld provide accelerometer signals alone. Because MoCA can be applied with either accelerometer or gyroscope inputs, or both, it remains compatible across both types of datasets.
Performance is evaluated under both linear probing (LP) and fine-tuning (FT), and results are compared with and without augmentation.

Subject heterogeneity can also impact model generalization, as differences in age, physical condition, and lifestyle can substantially influence movement patterns. To account for this variability, we conduct subject-wise evaluations following the approach in \cite{harnet}: five-fold cross-validation for datasets with more than five participants and leave-one-subject-out validation for datasets with five or fewer subjects. This evaluation approach prevents any overlap between training and test subjects and provides a systematic assessment of models' robustness to both sensor variability and unseen subjects.
Results are summarized as mean top-1 classification accuracy with standard deviation across folds.

We compare against three baselines to isolate the contributions of MoCA's masking scheme, architecture, and pre-training strategy. \textbf{MAE Synchronized Masking} uses the same ViT backbone as MoCA but applies synchronized masking where all modalities are masked/unmasked together, rather than MoCA's cross-modality masking approach (Section~\ref{discriminative_task}). \textbf{Yuan et al.} \cite{harnet} represents an alternative self-supervised learning framework using multi-task pre-training on accelerometer data with axis-invariant augmentations. Finally, \textbf{Supervised ViT-Base} provides a standard fully supervised baseline trained from random initialization, consistent with our evaluation in Section~\ref{discriminative_task}.


\begin{table*}[tb]
\centering
\caption{\textbf{Transfer learning performance across six activity recognition datasets.} 
Models pre-trained on UCI-HAR are evaluated on WISDM, IMWSHA, OPPORTUNITY, ADL, PAMAP2, and RealWorld. 
Top-1 classification accuracy (mean $\pm$ standard deviation) is reported under linear probing (LP) and fine-tuning (FT). 
MoCA is compared with MAE Synchronized Masking, Yuan et al. \cite{harnet} (pre-trained on UCI-HAR), and a fully supervised ViT-Base \cite{vit}. 
Gray shading indicates the best performance in each dataset.}
\resizebox{\textwidth}{!}{
\begin{tabular}{@{}llcccccc@{}}
\toprule
Model & Setting & WISDM & IMWSHA & OPPORTUNITY & ADL & PAMAP2 & RealWorld \\
\midrule
MAE Synchronized Masking (w/o aug.) & LP & 61.42 $\pm$ 4.19 & 50.48 $\pm$ 7.03 & 46.06 $\pm$ 18.69 & 70.61 $\pm$ 6.83 & 65.61 $\pm$ 3.72 & 63.98 $\pm$ 6.36 \\
MAE Synchronized Masking (w/ aug.) & LP & 65.65 $\pm$ 4.35 & 62.95 $\pm$ 6.47 & 47.65 $\pm$ 12.35 & 71.62 $\pm$ 3.05 & 72.83 $\pm$ 2.00 & 66.18 $\pm$ 6.36 \\
Yuan et al. \cite{harnet} (pre-trained on UCI-HAR) & LP & 59.01 $\pm$ 4.81 & 39.95 $\pm$ 5.34 & \cellcolor{gray!30} \textbf{59.31 $\pm$ 8.40} & 49.04 $\pm$ 1.40 & 56.40 $\pm$ 3.85 & 65.56 $\pm$ 4.50 \\
MoCA (w/o aug.) & LP & 66.08 $\pm$ 4.49 & 65.18 $\pm$ 4.61 & 53.34 $\pm$ 13.82 & 68.48 $\pm$ 2.80 & 68.74 $\pm$ 2.12 & 68.85 $\pm$ 3.89 \\
\textbf{MoCA (w/ aug.)} & LP & \cellcolor{gray!30} \textbf{68.98 $\pm$ 4.76} & \cellcolor{gray!30} \textbf{72.27 $\pm$ 10.55} & 54.04 $\pm$ 16.60 & \cellcolor{gray!30} \textbf{74.15 $\pm$ 0.74} & \cellcolor{gray!30} \textbf{77.80 $\pm$ 4.13} & \cellcolor{gray!30} \textbf{71.54 $\pm$ 4.14} \\
\cmidrule{1-8}
MAE Synchronized Masking (w/o aug.) & FT & 66.33 $\pm$ 4.31 & 68.39 $\pm$ 5.87 & 58.19 $\pm$ 14.78 & 75.30 $\pm$ 1.88 & 73.63 $\pm$ 1.92 & 73.19 $\pm$ 5.40 \\
MAE Synchronized Masking (w/ aug.) & FT & 71.54 $\pm$ 4.19 & 76.69 $\pm$ 10.55 & 67.03 $\pm$ 5.38 & 75.75 $\pm$ 2.46 & 75.00 $\pm$ 2.12 & 71.88 $\pm$ 4.06 \\
Yuan et al. \cite{harnet} (pre-trained on UCI-HAR) & FT & \cellcolor{gray!30} \textbf{76.90 $\pm$ 4.46} & 78.62 $\pm$ 10.02 & 59.34 $\pm$ 17.80 & \cellcolor{gray!30} \textbf{88.62 $\pm$ 1.49} & 76.75 $\pm$ 4.38 & \cellcolor{gray!30}\textbf{78.63 $\pm$ 6.39}  \\
MoCA (w/o aug.) & FT & 70.51 $\pm$ 4.53 & 73.49 $\pm$ 8.52 & 66.80 $\pm$ 4.02 & 77.61 $\pm$ 1.88 & 73.02 $\pm$ 2.66 & 72.93 $\pm$ 3.83 \\
\textbf{MoCA (w/ aug.)} & FT & 73.88 $\pm$ 4.40 & \cellcolor{gray!30} \textbf{80.54 $\pm$ 9.39} & \cellcolor{gray!30} \textbf{67.17 $\pm$ 5.97} & 81.64 $\pm$ 1.68 & \cellcolor{gray!30} \textbf{79.90 $\pm$ 3.99} & 77.43 $\pm$ 3.22 \\
\cmidrule{1-8}
ViT-Base \cite{vit} & / & 60.54 $\pm$ 4.55 & 62.31 $\pm$ 7.88 & 60.05 $\pm$ 8.35 & 71.55 $\pm$ 2.85 & 71.75 $\pm$ 5.14 & 71.39 $\pm$ 4.55 \\
\bottomrule
\end{tabular}
}
\label{tab:kfold}
\end{table*}

Table~\ref{tab:kfold} reports transfer learning performance on six benchmark datasets under linear probing (LP) and fine-tuning (FT), with results given as mean top-1 classification accuracy ($\pm$ standard deviation) across subject-wise cross-validation folds.
MoCA with augmentation achieves the best overall transfer learning performance, yielding the highest average accuracies under both LP and FT among all approaches. Under LP, it achieves the top accuracy on five of the six datasets and improves the average accuracy to 69.8\%, a 14.9 percentage point gain over Yuan et al. \cite{harnet}. Under FT, MoCA leads on three of the six datasets and attains the highest overall average accuracy of 76.76\%, slightly surpassing Yuan et al. (76.48\%).
Yuan et al. achieves competitive FT performance on some datasets (e.g., WISDM, ADL) but lags overall, particularly in LP task where transferable representations are critical. 
MoCA consistently outperforms MAE Synchronized Masking under both LP and FT, confirming the benefit of cross-modality masking. 
The supervised ViT-Base performs the worst across datasets, underscoring the value of SSL pre-training for generalization.

Notably, MoCA transfers effectively from six-channel pre-training (accelerometer + gyroscope) to accelerometer-only downstream datasets (WISDM, RealWorld, OPPORTUNITY, PAMAP2, ADL), demonstrating robust generalization across input dimensionalities. This cross-modality flexibility is particularly relevant in clinical contexts, where sensor configurations often vary.

The evaluation also shows that augmentation plays a critical role. Both MoCA and MAE Synchronized Masking show clear drops in accuracy without augmentation, consistent with prior reports on the importance of large and diverse training corpora for self-supervised learning \cite{mae}. Our augmentation strategy, tailored to wearable health data, alleviates this limitation and strengthens transfer performance.

\begin{figure*}[tb]
    \centering
    \includegraphics[width=\linewidth]{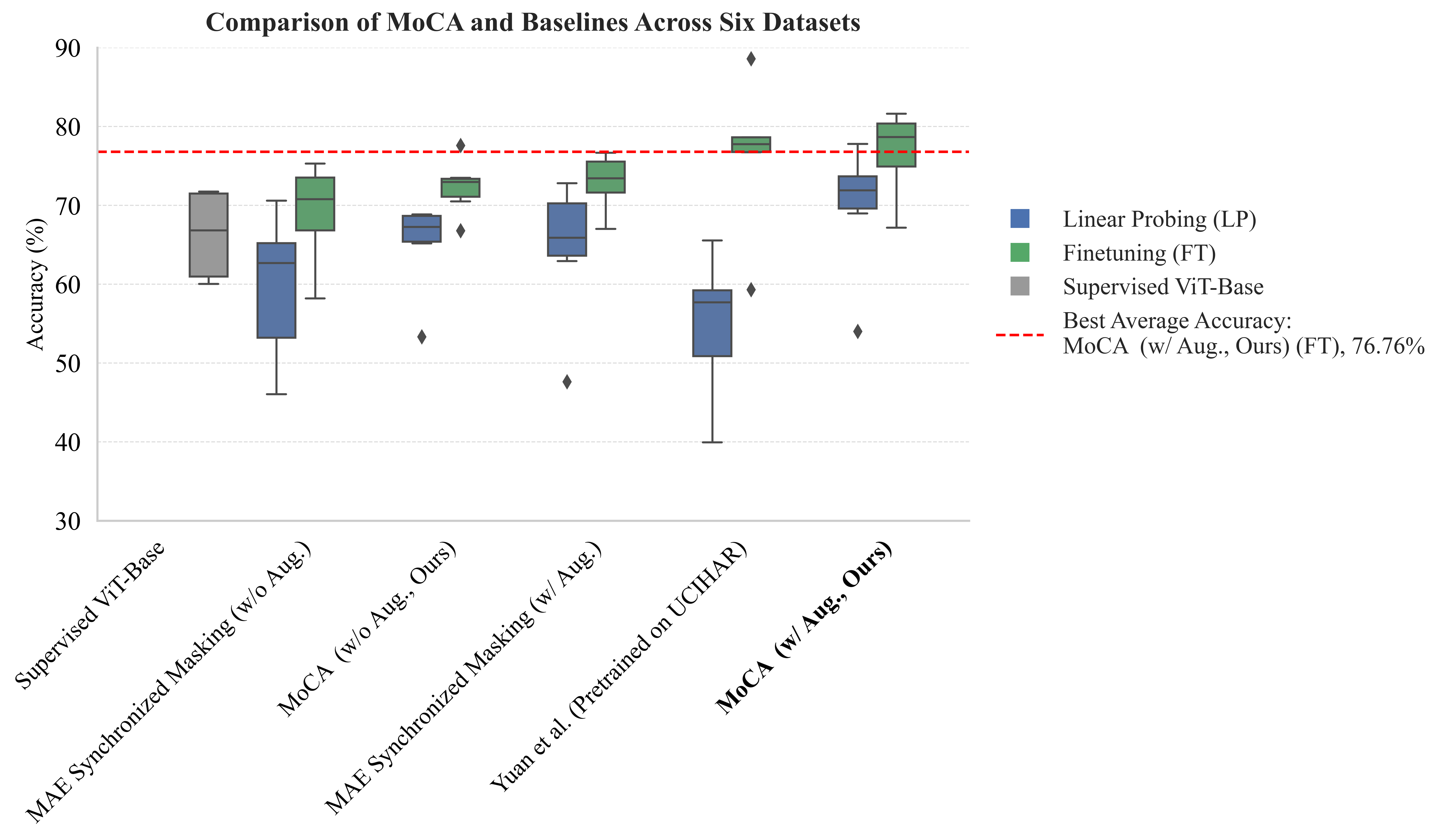}
    \caption{\textbf{Comparison of model performance across six datasets under linear probing and fine-tuning.} Boxplots show the distribution of top-1 classification accuracies across six benchmark datasets for each model, evaluated with and without data augmentation. Each point within a boxplot represents the mean test accuracy (\%) for a single dataset, obtained from subject-wise cross-validation. }
    \label{fig:transfer-boxplot}
\end{figure*}

Further insights are illustrated in the boxplot (Figure~\ref{fig:transfer-boxplot}), which summarizes the distribution of LP and FT performance across datasets. The performance margin is particularly evident under LP, where the frozen pre-trained encoder is used. This indicates that MoCA’s masking strategy and pre-training design yield stronger representations immediately useful for downstream tasks. In contrast, the narrower gap under FT suggests that weaker initial representations from competing models can be partially compensated through extensive parameter updates.  

Together, these findings demonstrate that MoCA provides the most robust and generalizable representations across diverse datasets and evaluation protocols. Its strong LP and FT performance, combined with the benefits of augmentation, highlights its practical value for wearable device data in clinical applications, where labeled data are scarce and sensor configurations are heterogeneous.

\subsection{Generative Task}
\label{generative_task}

Missing data are a common challenge in wearable device data analytics, particularly in clinical and free-living scenarios where signal loss may occur due to sensor glitches, battery depletion, device malfunction, or participant non-compliance. These issues are magnified in multi-sensor settings, where the absence of one modality can compromise downstream analyses. 
Motivated by this challenge, we assess each model’s ability to reconstruct missing data on the UCI-HAR test set across four representative tasks: (i) \textbf{Random Imputation}, which tests local reconstruction by randomly masking small patches; (ii) \textbf{Temporal Imputation}, which evaluates long-term dependency modeling by masking all data over consecutive intervals; (iii) \textbf{Sensor Imputation}, which measures cross-modality learning by masking an entire sensor channel; and (iv) \textbf{Temporal Extrapolation}, which requires forecasting future values by masking all data beyond a certain time point. A 70\% masking ratio is applied in all tasks except sensor imputation.  

\begin{figure*}[tb]
    \centering
    \includegraphics[width=\textwidth]{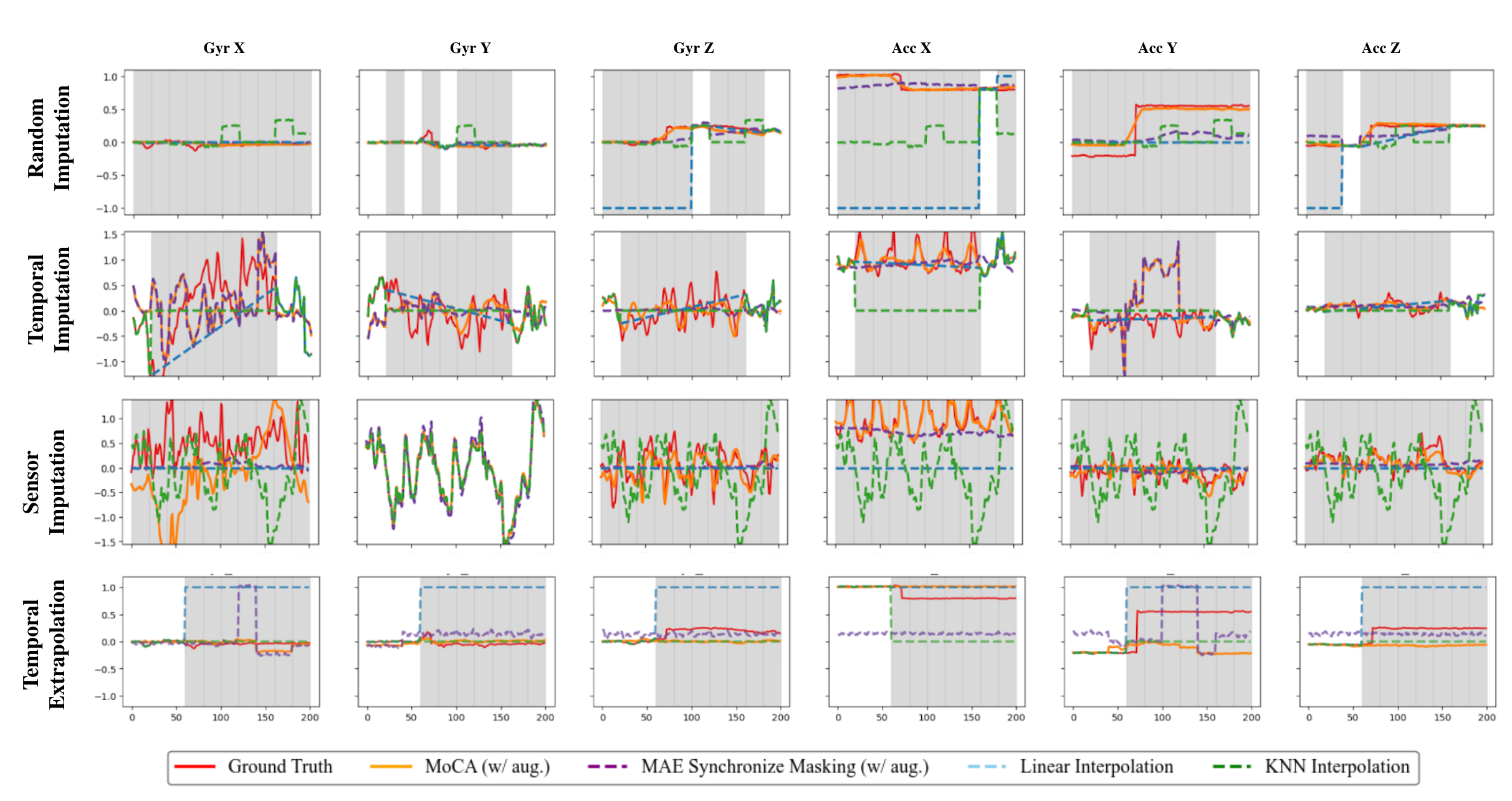}
    \caption{\textbf{Visualization of missing data imputation on the UCI-HAR test set.} Randomly selected representative reconstructed sequences are shown for the four tasks: random imputation, temporal imputation, sensor imputation, and temporal extrapolation.}
\label{fig:imputation_methods}
\end{figure*}

We benchmark MoCA against both statistical and self-supervised baselines. Traditional methods include (1) linear interpolation, (2) nearest-neighbor imputation, and (3) multivariate imputation by chained equations (MICE) \cite{mice}. We also compare against MAE Synchronized Masking (with and without augmentation) as a self-supervised baseline.
The supervised ViT-Base and Yuan et al. baselines are not included here, as neither model is designed to perform generative reconstruction of missing data. 
Following the Large Sensor Model (LSM) protocol \cite{genrative_baseline}, evaluation is performed using mean squared error (MSE) and mean absolute error (MAE) between the predicted missing value and the ground truth value, where lower values indicate better reconstruction. We found that MICE and nearest neighbor methods both degrade under high masking ratios and often produce visually similar outputs. To avoid redundancy, MICE is omitted from the visualizations (Figure \ref{fig:imputation_methods}) but reported in Table~\ref{tab:generative_task}.

\begin{table*}[t]
    \centering
    \caption{\textbf{Imputation performance on the UCI-HAR test set.} Models are evaluated on four tasks reflecting common missing data patterns: random imputation (random patches removed across time and modalities), temporal imputation (consecutive intervals removed across all modalities), sensor imputation (all intervals from one modality removed), and temporal extrapolation (final intervals across all modalities removed). Missing inputs are reconstructed using the pre-trained encoder–decoder, and accuracy is quantified by mean absolute error (MAE) and mean squared error (MSE), with lower values indicating better performance. Gray shading highlights the best performance in each task. All models are pre-trained on UCI-HAR; a 70\% missing ratio is applied in all tasks except sensor imputation. For sensor imputation, only one modality is kept visible, and the model reconstructs all other modalities.}
    \resizebox{\linewidth}{!}{
    \begin{tabular}{lcc|cc|cc|cc}
    \toprule
    \textbf{Model} / \textbf{Error (MAE / MSE) $\downarrow$}
    & \multicolumn{2}{c|}{\textbf{Random Imputation}}  
    & \multicolumn{2}{c}{\textbf{Temporal Imputation}} 
    & \multicolumn{2}{c}{\textbf{Sensor Imputation}} 
    & \multicolumn{2}{c}{\textbf{Temporal Extrapolation}} \\
    \midrule
    Linear Interpolation & 0.587 & 0.689  & 0.546 & 0.634  & 0.252 & 0.189 & 0.600 & 0.635 \\ 
    Nearest Neighbor Imputation & 0.341 & 0.284  & 0.240 & 0.180  & 0.412 & 0.377 & 0.211 & 0.158\\ 
    MICE              & 0.289 & 0.251  & 0.410 & 0.375  & 0.210 & 0.158 & 0.211 & 0.158\\ 
    \midrule
    MAE Synchronized Masking (w/o aug.) & 0.275 & 0.162  & 0.262 & 0.191  & 0.288 & 0.170 & 0.266 & 0.194 \\
    MAE Synchronized Masking (w/ aug.) & 0.380 & 0.281  & 0.306 & 0.231  & 0.494 & 0.459 & 0.180 & 0.122 \\
    \midrule
    MoCA (w/o aug., Ours) & 0.086 & 0.043  & \cellcolor{gray!30}\textbf{0.112} & \cellcolor{gray!30}\textbf{0.059}  & 0.160 & 0.089 & \cellcolor{gray!30}\textbf{0.111} & \cellcolor{gray!30}\textbf{0.059}\\
    \textbf{MoCA (w/ aug., Ours)} & \cellcolor{gray!30}\textbf{0.076} & \cellcolor{gray!30}\textbf{0.027}  & 0.153 & 0.102  & \cellcolor{gray!30}\textbf{0.140} & \cellcolor{gray!30}\textbf{0.063} & 0.177 & 0.121 \\
    \bottomrule
    \end{tabular}
    }
    \label{tab:generative_task}
\end{table*}


As shown in Table~\ref{tab:generative_task}, MoCA substantially outperforms all statistical baselines and the MAE Synchronized Masking model across most tasks. In random and sensor imputation, MoCA with augmentation achieves the lowest reconstruction error, demonstrating its ability to leverage cross-modality dependencies. In temporal imputation and temporal extrapolation, MoCA without augmentation performs better than with augmentation, suggesting that augmentation-induced domain shifts may reduce sensitivity to authentic temporal dependencies. Notably, in the temporal imputation task where large contiguous segments are masked, linear interpolation, nearest neighbor imputation and MICE fail to recover meaningful patterns, whereas MoCA uses cross-modality learning to collect information from scattered unmasked patches across time and modalities to reconstruct the entire segment.

Figure~\ref{fig:imputation_methods} provides qualitative examples of reconstructed sequences. In the random imputation task, MoCA uses only a few visible patches scattered across modalities to faithfully reconstruct the entire series. In sensor imputation, when all channels except a single modality (e.g., Gyr Y) are masked, traditional methods collapse with predicting flat or zero-valued signals, whereas MoCA leverages the remaining input to reconstruct realistic temporal patterns across all sensors. These qualitative results reinforce MoCA’s ability to capture both intra- and inter-modality temporal structure and to integrate information across modalities.

These results above highlight MoCA’s strength as a generative model for multi-modal wearable device data. It addresses the critical challenge of reliable data recovery in clinical and free-living monitoring by producing realistic reconstructions under diverse missingness patterns.


\subsection{Ablation Studies}

\begin{figure}[tb]
    \centering
    \includegraphics[width=1\linewidth]{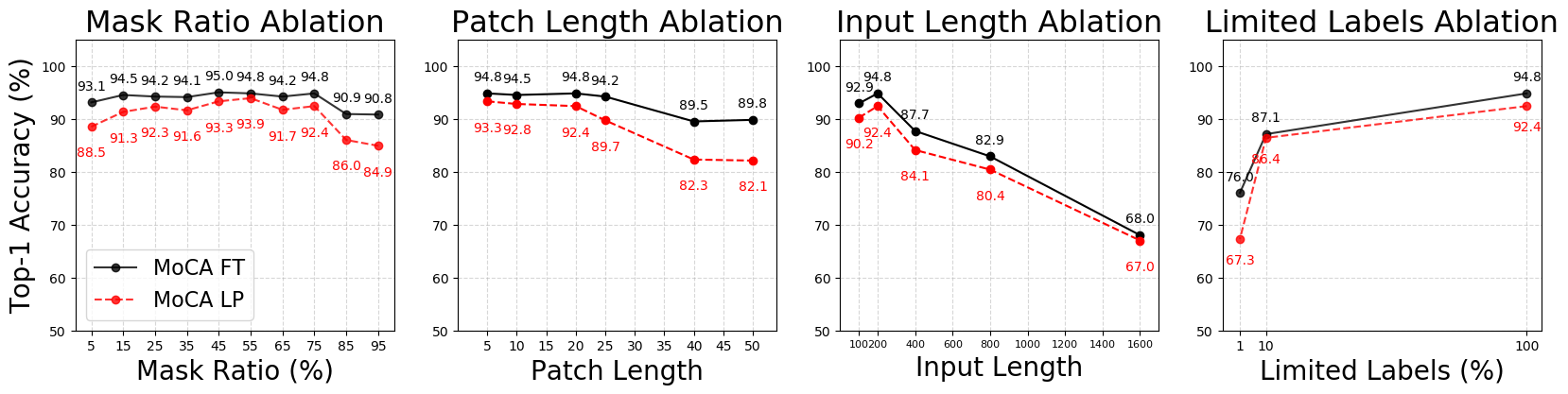}
    \caption{\textbf{MoCA ablation study results.} Fine-tuning (FT) and linear probing (LP) top-1 accuracies on UCI-HAR under varying mask ratios, patch lengths, input lengths, and label fractions. All experiments use MoCA (w/o augmentation) with a ViT-Base encoder pretrained for 4{,}000 epochs. }
    \label{fig:ablation}
\end{figure}

We conduct ablation studies to isolate the contribution of individual design choices to MoCA’s performance: \emph{input shape} (results in Table~\ref{tab:inputsize}), hyperparameters including \emph{mask ratio}, \emph{patch length}, \emph{input length}, and \emph{limited labels} (results in Figure~\ref{fig:ablation}). Unless noted otherwise, all ablations are run without augmentation. Training settings match MoCA (w/o aug.): pretraining for 4{,}000 epochs and downstream fine-tuning/linear probing for 50 epochs with identical architecture and optimizer.

\vspace{3mm}
\noindent \textbf{Input Shape}
\vspace{0.5mm}

To disentangle the effect of input shape from the masking strategy, we compare MoCA with several synchronized masking MAE variants that differ only in how the six sensor channels are arranged as input. We illustrate input dimensions as [row, column, channel]. All baseline models are variants of MAE with a synchronized masking scheme, meaning the same time intervals are masked across all modalities. The difference lies only in how the six sensor channels are arranged as input:  
1) \textbf{MAE Original} (\texttt{[1, 200, 6]}): a naive extension of the original MAE, treating the six modalities as channels;
2) \textbf{MAE Stacked Inputs} (\texttt{[6, 200, 6]}): six inputs, each with six channels, are stacked along the row dimension while preserving the original channel dimension;
3) \textbf{MAE Synchronized Masking} (\texttt{[6, 200, 1]}): six channels are concatenated into rows, matching the input shape used by MoCA.  
For fairness, the training schedule is adjusted so that the total number of iterations remains constant despite changes in sample size caused by stacking. 

Table~\ref{tab:inputsize} summarizes results. Across synchronized masking variants with different input shapes, MoCA (w/o aug.) consistently achieves the highest accuracy, with 2.1 percentage points higher in FT (94.8 vs.\ 92.7) and 2.0 percentage points higher in LP (92.4 vs.\ 90.4) over the best baseline, indicating that MoCA’s gains arise from cross-modality masking rather than trivial input reshaping.

\begin{table}[t]
\centering
\scriptsize
\setlength{\tabcolsep}{4pt}
\renewcommand{\arraystretch}{0.9}
\caption{\textbf{Input shape ablation.} The top-1 fine-tuning (FT) and linear probing (LP) accuracies of the models of different input shapes are tested on UCI-HAR.}
\begin{tabular}{l|c|c}
\toprule
\textbf{Model (input shape)} & \textbf{FT Top-1 (\%)} & \textbf{LP Top-1 (\%)} \\
\midrule
MAE Original (\texttt{[1, 200, 6]}) & 92.70 & 90.40 \\ 
MAE Stacked Inputs (\texttt{[6, 200, 6]}) & 85.60   & 50.80 \\ 
MAE Synchronized Masking (\texttt{[6, 200, 1]}) & 91.70 & 86.30  \\ 
\midrule
\textbf{MoCA (w/o aug., Ours)} (\texttt{[6, 200, 1]}) &  \cellcolor{gray!30} \textbf{94.80} & \cellcolor{gray!30} \textbf{92.40} \\
\bottomrule
\end{tabular}
\label{tab:inputsize}
\end{table}

\vspace{8mm}
\noindent \textbf{Mask Ratio}
\vspace{0.5mm}

We vary the masking ratio from 5\% to 95\% in 5\% increments. MoCA achieves stronger performance with ratios 45\%, 55\% and 75\% with 94.8\% to 95.0\% accuracy for FT and 92.4\% to 93.9\% accuracy for LP. These results align with findings for image-based MAE model \cite{mae}, which reported optimal performance near 75\% masking. Top-1 accuracy drops down quickly once the ratio exceeds 75\% for both FT and LP. Still, MoCA demonstrates robust classification performance with mask ratio at 95\% with 90.8\% accuracy for FT and 84.9\% accuracy for LP.

\vspace{3mm}
\noindent \textbf{Patch Length}
\vspace{0.5mm}

We evaluate the effect of varying the patch length $L_p \in \{5, 10, 20, 25, 40, 50\}$ while fixing the input length at 200, which corresponds to a 4-second time window. Since there are six sensor axes, the number of patches per input is $(200 / L_p) \times 6$, resulting in 240, 120, 60, 48, 30, and 24 patches, respectively. Shorter patches $\{5, 10, 20\}$ yield the best results: FT top-1 accuracies of 94.8\%, 94.5\%, and 94.8\%, and LP top-1 accuracies of 93.3\%, 92.8\%, and 92.4\%. Performance degrades as patch length increases. A possible reason is that, under a fixed mask ratio, larger patches correspond to longer contiguous spans of data being removed, leaving fewer continuous visible segments. This reduces the model’s ability to capture fine-grained local temporal patterns across modalities.



\vspace{3mm}
\noindent \textbf{Input Length}
\vspace{0.5mm}

We vary the input length $L \in \{100, 200, 400, 800, 1600\}$ for each modality while fixing the number of patches at 10 per modality. This yields corresponding patch lengths of $L_p = \{10, 20, 40, 80, 160\}$.
The best performance is achieved at $L=200$, which corresponds to a 4-second window, while both FT and LP top-1 accuracies decline for input lengths greater than 200. This shows that in the context of physical movements captured by the UCI-HAR dataset, using a 4-second input contributes to creating the best learned representation during pre-training.
Another factor to consider is that because the number of patches in each input is fixed, increasing $L$ forces each patch to span a longer segment of the signal, effectively increasing $L_p$.
As in the patch length ablation study, longer patches reduce the amount of fine-grained visible data, which limits the model’s ability to learn detailed temporal patterns across modalities.

\vspace{3mm}
\noindent \textbf{Limited Labels}
\vspace{0.5mm}

We evaluate the model performance under limited label conditions for downstream classification, following the protocol in \cite{swav}. FT and LP are conducted on subsets of UCI-HAR with $1\%$ and $10\%$ of the labels, using pre-trained MoCA. These subsets are sampled to preserve the overall label distribution of the entire training set.
MoCA maintains stable classification accuracy well above the random-guess baseline of $1/7=14.3\%$. With 10\% of the labeled data (approximately 185 inputs), FT and LP achieve 87.1\% and 86.4\% accuracy respectively. With only 1\% of the labeled data (about 19 inputs), accuracy remains at 76.0\% for FT and 67.3\% for LP. These results demonstrate that MoCA’s pre-trained representations are robust to label scarcity: downstream accuracies remain at a reasonable level even in extreme low-label settings.


\subsection{Qualitative Analysis and Latent Space Visualization}
To complement the quantitative results, we inspect the learned representations through qualitative analyses and interpretable visualizations.
Specifically, we use UMAP to visualize the latent space structure and attention maps to highlight which signal patches contribute most to MoCA’s activity classifications.

\begin{figure}[h]
    \centering
    \includegraphics[width=\textwidth, trim=0 0 0 0, clip]{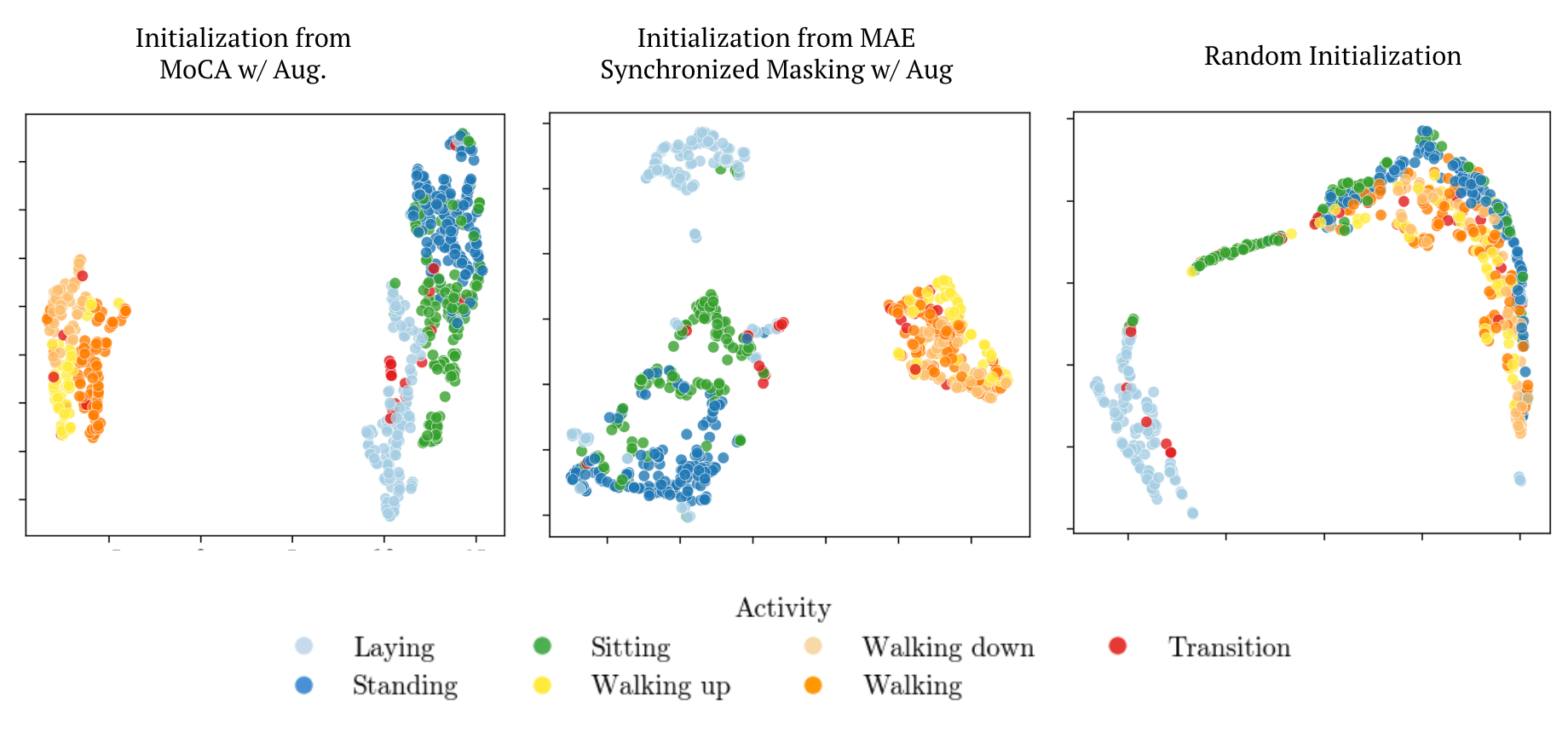}
    \caption{\textbf{UMAP visualizations of class token embeddings from three initializations.} All models use the ViT encoder for fine-tuning, but differ in weight initialization: initializing from MoCA with augmentation pre-training (left), initializing from MAE Synchronized Masking with augmentation pre-training (middle), and random initialization (right).}
    \label{fig:umap}
\end{figure}

\vspace{6mm}
\noindent \textbf{UMAP}
\vspace{0.5mm}

Figure \ref{fig:umap} shows Uniform Manifold Approximation and Projection (UMAP) \cite{umap} visualizations of class token embeddings from the UCI-HAR test set. We compare three fine-tuned models that share the same ViT backbone but differ in initialization: MoCA with augmentation, MAE with synchronized masking and augmentation, and random initialization. 
The model initialized from MoCA (Figure \ref{fig:umap}, left) produces well-separated clusters for all activity classes, including closely related postures such as sitting versus standing and the three walking variants (walking, walking downstairs, walking upstairs). This separation is driven by MoCA’s cross-modal masking scheme, which encourages the model to align and reconstruct signals across modalities, yielding representations that capture subtle distinctions between such fine-grained activities. 
In contrast, the model initialized from MAE with synchronized masking (Figure \ref{fig:umap}, middle) fails to capture cross-modal dependencies and merges the fine-grained walking activities. 
The randomly initialized model (Figure \ref{fig:umap}, right) performs the worst, showing extensive cluster mixing due to the lack of informative priors and sole reliance on the limited supervised labels.
These results demonstrate that performance gains come from MoCA’s cross-modal representation learning rather than model capacity, consistent with the main results (Table \ref{tab:main}), and confirm that MoCA pre-training improves the quality of learned representations.

\begin{figure*}[t]
    \centering
    \includegraphics[width=0.85\linewidth]{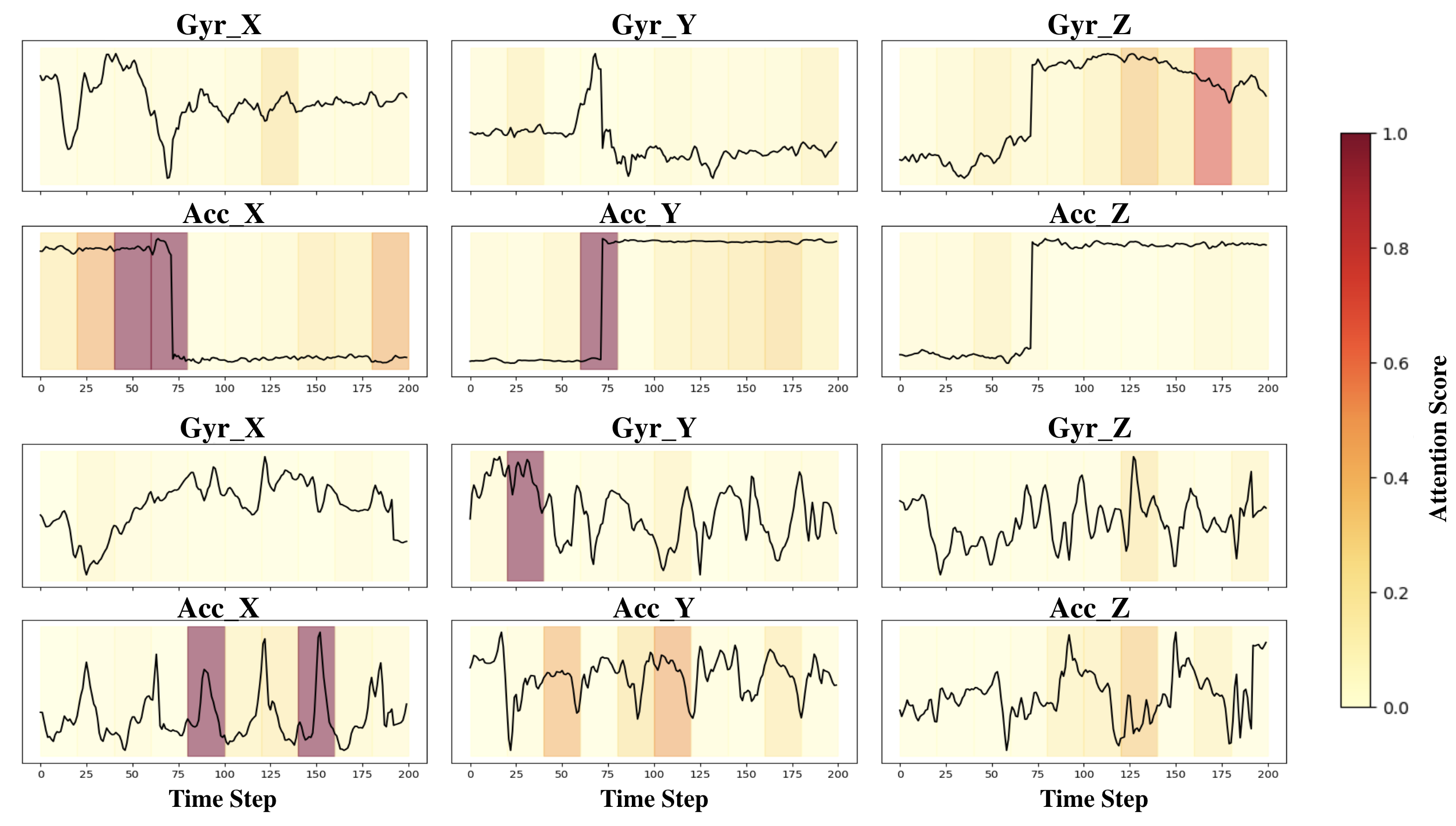}
    \caption{\textbf{Attention maps from MoCA (w/ aug.) for two example activities.} Top: \textit{transition}. Bottom: \textit{walking up}. Patch colors indicate attention scores, with yellow for low and dark red for high attention. Highly attended patches contribute more to the final classification of activity. 
    Black lines denote the raw input signals. Each input spans 4 seconds, divided into patches of 0.4 seconds each. All six sensor axes (gyroscope X/Y/Z and accelerometer X/Y/Z) are shown for every input.
    }
    \label{fig:overall_attention_maps}
    \vspace{-5mm}
\end{figure*}

\vspace{2mm}
\noindent \textbf{Attention Maps}
\vspace{0.5mm}

To provide interpretable insights into how MoCA makes predictions, Figure~\ref{fig:overall_attention_maps} overlays attention scores on the raw input signals for two example activities: \textit{transition} (top) and \textit{walking up} (bottom). Each input spans 4 seconds and includes all six sensor axes (gyroscope X/Y/Z and accelerometer X/Y/Z), segmented into patches of 0.4 seconds. Dark red patches indicate regions with high attention, which contribute most to the final classification. For \textit{transition}, attention concentrates around abrupt signal changes, while for \textit{walking up}, it highlights the periodic motion pattern. These patterns show that MoCA attends to semantically meaningful segments that align closely with the underlying activity labels.



\section{Methods}
\label{proposed_method}

MoCA is a self-supervised masked autoencoder framework for multi-modal wearable time series. It extends the MAE with a cross-modality masking scheme, targeted augmentation, and reconstruction-based pre-training, followed by fine-tuning and linear probing for downstream evaluation.
Below we provide technical details of the proposed framework.

\subsection{Dataset}
To support reproducibility and clarity, we provide detailed descriptions of all public datasets used in our study. A summary is provided in Table \ref{tab:dataset_summary}, Appendix \ref{dataset}. 

\subsection{Model Architecture}
\label{model_arch}

MoCA builds on the Masked Autoencoder (MAE) framework \cite{mae}, adapting it for multivariate time series through modified patch embedding. Like the original MAE, it follows an autoencoding structure with a Transformer-based encoder that maps observed patches to latent representations and a decoder that reconstructs the masked patches. 
Both encoder and decoder use the Vision Transformer (ViT) architecture \cite{vit}. The pre-trained model is then evaluated on downstream classification tasks via fine-tuning and linear probing. Detailed components are described below and illustrated in Figure~\ref{fig:fig1}.


\vspace{3mm}
\noindent \textbf{Time Series Input}
\vspace{0.5mm}

We adapt MAE to multivariate time series by representing each sample as $x \in \mathbb{R}^{C \times L}$, where $C$ denotes the sensor modalities (e.g., accelerometer and gyroscope axes) and $L$ the per-modality sequence length.
We segment each modality into non-overlapping windows of length $L_p$, producing 
$P = \lfloor L / L_p \rfloor$ patches per modality and a total of $C \times P$ patches per input. Thus, the tokenized input has shape 
$\mathbb{R}^{C \times P \times L_p}$. Each patch is then linearly projected into an embedding space of dimension $D$ before being passed to the encoder.

\vspace{3mm}
\noindent \textbf{Encoder and Decoder}
\vspace{0.5mm}

MoCA uses ViT-Base (ViT-B), a stack of 12-layer standard Transformer blocks \cite{transformer}, as its encoder.
With a masking ratio of 0.75, only the $0.25 \cdot P \cdot C$ unmasked patches are fed into the Transformer blocks. The decoder is also composed of standard Transformer blocks of fewer layers (8-layer), following the asymmetric architecture of MAE \cite{mae}. 
The encoded patches are first padded with trainable masked tokens in their original position. 
Fixed 2D sinusoidal positional embeddings are then added to provide information of their respective modality and temporal locations. 
A linear reconstruction head on top of the decoder stack outputs the predicted values for all patches in the input time series.

\vspace{3mm}
\noindent \textbf{Objective}
\vspace{0.5mm}

The MAE decoder learns to reconstruct the input multivariate series by predicting the values in each data point. Our training objective is to reduce the loss in the form of mean squared error (MSE) between the reconstructed and the original input, averaged over \textbf{all} patches.

\subsection{Masking Schemes}


Given a mask ratio $\rho$, a proportion $\rho$ of tokenized patches is randomly masked before being passed to the model. The encoder processes only the unmasked patches to generate latent representations, while the decoder reconstructs the missing patches. The choice of masking strategy is therefore critical, as it determines what contextual information the model must rely on. We evaluate two approaches:

\vspace{3mm}
\noindent \textbf{Cross-modality Masking (for MoCA)}
\vspace{0.5mm}

In MoCA, patches from different modalities are masked \textbf{independently} across time windows. This masking scheme provides more flexibility and allows the model to infer missing signals in one modality from the remaining signals in other modalities at the same or nearby time points. Such cross-modality dependencies are common in wearable device data (e.g., accelerometer and gyroscope streams often exhibit correlated but complementary patterns). Cross-modality masking encourages the encoder to learn richer representations that preserve both within-modality structure and cross-modality relationships critical for downstream tasks.



\vspace{3mm}
\noindent \textbf{Synchronized Masking}

\vspace{0.5mm}

As a comparison, synchronized masking removes the same time windows across \textbf{all} modalities simultaneously.
Several previous studies on MAE for multivariate time series \cite{li2023ti, zhang2023trid} adopt this masking scheme, treating different modalities, including separate axes of the same sensor, as channels analogous to the RGB channels in images. 
While straightforward to adapt from vision MAE, synchronized masking constrains the model by removing all modality information at a given time point all at once, limiting its ability to leverage cross-modal learning for reconstruction. As a result, it encourages the model to rely more on temporal correlations rather than inter-modality relationships.



\subsection{Data Augmentation}
   
\label{data_aug_sec}

\begin{algorithm}[t]
   \caption{Mixup-based augmentation for multivariate time series}
   \label{alg:augment_time_series}
\begin{algorithmic}
   \Statex \textit{Notation:} $\mathcal{U}(a,b)$ is the discrete uniform distribution over integers in $[a,b]$.
   \State \textbf{Input:} Dataset $\mathcal{X}$ containing sequences $x \in \mathbb{R}^{C \times L}$.
   \State \textbf{Output:} Augmented sequence $\tilde{x} \in \mathbb{R}^{C \times L}$.
   \State 1. Sample $x_1, x_2 \sim \mathcal{X}$ \Comment{same $C$ and $L$}
   \State 2. Sample segment length $\lambda \sim \mathcal{U}\!\big(\lceil 0.2L\rceil,\ \lfloor 0.5L\rfloor\big)$
   \State 3. Sample start indices $s_1, s_2 \sim \mathcal{U}(0, L-\lambda)$
   \State 4. Let $e_1 \gets s_1+\lambda$, $e_2 \gets s_2+\lambda$
   \State 5. \textbf{Joint splice across all modalities:} $x_1[:, s_1\!:\!e_1] \gets x_2[:, s_2\!:\!e_2]$
   \State 6. \textbf{return} $\tilde{x} \gets x_1$
\end{algorithmic}
\end{algorithm}

To enhance MoCA's ability to learn robust cross-modality representations, particularly for capturing activity transitions, we employ a targeted data augmentation strategy during pre-training that generates synthetic multi-modal samples containing activity transitions. Our approach addresses the challenge that naturally occurring datasets often under-represent critical transitional periods (e.g., transitions between different activity states or physiological conditions), which are essential for learning generalizable cross-modal dependencies.

As described in Algorithm \ref{alg:augment_time_series}, we utilize a mixup-based augmentation \cite{zhang2017mixup}.
The augmentation samples two sequences, extracts a random segment (20–50\% of the total length) from one, and splices it into another at a matched temporal position.
Because the spliced segment may not align smoothly with the surrounding context, the model cannot simply interpolate within a single modality; instead, it must rely on information from other modalities at the same time point to resolve inconsistencies. This repeated exposure to mismatched or corrupted local context during training acts as a form of structured noise injection. By training under these controlled disruptions, the model develops robustness to incomplete or corrupted inputs and strengthens its ability to generalize to unseen dynamics.
The augmentation can be applied either offline, where augmented samples are generated once and saved to disk for reproducibility, or on-the-fly during training to introduce additional diversity.

Under the MAE masking objective, these transitions require the decoder to reconstruct data, including those representing the transitions, using both temporal context and complementary modalities. This encourages representations that are robust to free-living activity transitions and mild discontinuities (e.g., brief posture changes or sensor micro-shifts). 

\section{Theoretical Analysis}
In this section, we provide an outline of a theoretical analysis of the proposed method. We establish that the Masked Autoencoder (MAE) reconstruction task is fundamentally connected to the kernel canonical correlation analysis (KCCA). 
Intuitively, our theory shows that MAE reconstruction reduces (via KCCA) to maximizing the alignment between the aggregate unmasked view and masked view, which captures not only within-modality cross-time patch associations, but also cross-modal relationships. Therefore, masking schemes that strengthen these view-level correlations yield better reconstructions. This insight is critical for multi-modal wearable devices where different modalities (e.g., accelerometer, gyroscope) capture complementary aspects of the same physiological state. The complete theoretical analysis is provided in the Appendix Section \ref{sect: theory}.

\subsection{Core Theoretical Result}

Consider multi-modal time series data from wearable devices, where each window contains $C$ modalities (sensor types) divided into $P$ temporal patches. During MAE pre-training, we randomly mask a portion of these patches and train the model to reconstruct them from the unmasked remainder. 

It is shown in \cite{zhang2022mask} that under standard assumptions, the MAE loss is lower-bounded by the following alignment term:
$$L_{\text{MAE}} \geq -\mathbb{E}[\langle h(x^U), h_g(x^M) \rangle] + \text{const},$$
where $h = g \circ f$ represents the encoder-decoder composition, and $x^U$ and $x^M$ represent unmasked and masked views of the inputs, respectively, as described in Appendix \ref{sect: theory}. 
Based on our derivation, this bound is tight under the normalization assumption.

Our theoretical analysis reveals that given a fixed masking scheme,
maximizing this alignment term is equivalent to solving a kernel canonical correlation analysis (KCCA) problem:
$$(\alpha^*, \beta^*) = \arg\max_{\alpha, \beta} \frac{\alpha^{\top} K_U K_M \beta}{\sqrt{\alpha^{\top} K_U^{2} \alpha \cdot \beta^{\top} K_M^{2} \beta}},$$
where $K_U$ and $K_M$ are Gram matrices capturing similarities within unmasked and masked views in their corresponding Reproducing Kernel Hilbert Spaces. The optimal value of this KCCA objective determines how well the masking strategy preserves learnable within- and cross-modal structure.

In the special case where the raw data are first mapped into feature representations that capture semantics of the data, and linear kernels are used to measure similarities, the KCCA formulation reduces to the classical canonical correlation analysis (CCA):
$$\max_{w_u, w_m} \frac{w_u^{\top} \Sigma_{U\!M} w_m}{\sqrt{(w_u^{\top} \Sigma_{U\!U} w_u)(w_m^{\top} \Sigma_{M\!M} w_m)}},$$
where $\Sigma_{U\!M}$ is the cross-covariance between feature mappings of the unmasked and masked views, and $\Sigma_{U\!U}$ and $\Sigma_{M\!M}$ are covariances among features mappings of the unmasked and masked views, respectively.
The solution to the CCA problem is given by the top singular value $\sigma_1$ of the normalized cross-covariance matrix
$\Gamma = \Sigma_{U\!U}^{-1/2} \Sigma_{U\!M} \Sigma_{M\!M}^{-1/2}$.
This singular value $\sigma_1(\Gamma)$ is the key quantity that controls reconstruction quality: larger values reflect stronger associations between unmasked and masked views, which translate into better MAE reconstructions.

\subsection{Practical Implications for Masking Strategy Design}

This theoretical analysis explains the outstanding performance of the proposed MoCA masking strategy: a masking scheme that leads to a larger KCCA objective will result in a smaller lower bound for the MAE loss.
Intuitively, the maximum of the KCCA objective measures the strength of association, both within- and cross-modality, between the unmasked and masked views under a given masking scheme. 
In datasets with strong cross-modality correlation such as multi-sensor wearable data, this criterion highlights why some masking strategies succeed and others fall short. Synchronized masking, which removes entire time slices across all modalities, reduces the potential association between unmasked and masked views because no modality remains available to aid reconstruction at a given time point. In contrast, MoCA masks patches independently across modalities within each window, preserving cross-modal cues so unmasked modalities can aid reconstruction of masked ones.

\subsection{Empirical Validation}

We validated this theoretical insight through a simulation study using synthetic transition data designed to mimic scenarios such as walking-to-sitting transitions. One example sample in the synthetic data is shown in Figure \ref{simulation_data}. 
By computing $\sigma_1(\Gamma)$ under different masking schemes (details in Appendix \ref{sect: theory}), we observed higher alignment with MoCA masking ($\sigma_1 = 0.925$) compared to synchronized masking ($\sigma_1 = 0.835$). 
These results confirm that MoCA more effectively leverages the underlying structure of multi-sensor wearable data, where diverse sensors capture complementary facets of human activity and physiology.

\begin{figure}[h]
\centering
\includegraphics[width=0.55\linewidth]{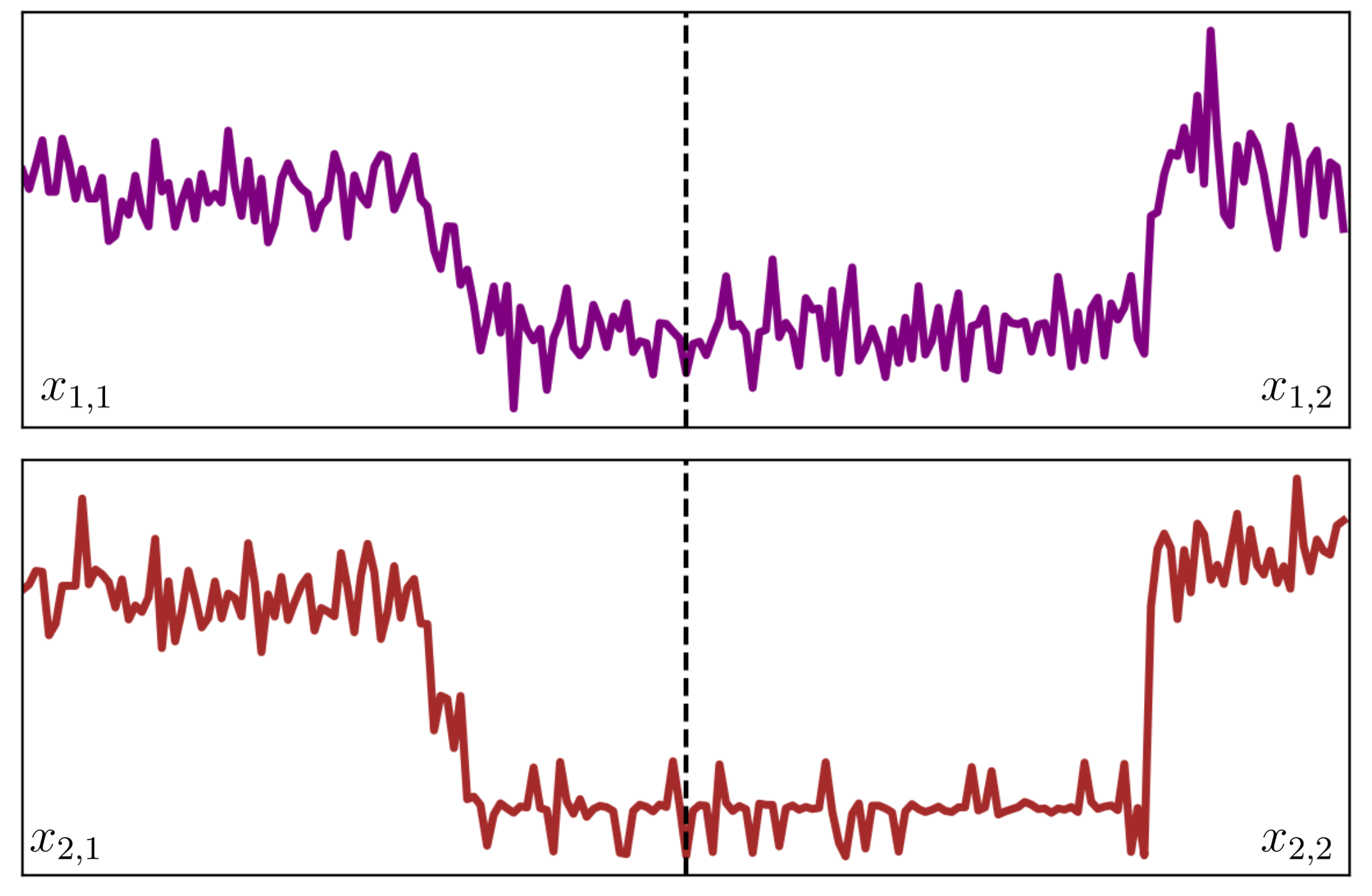}
\caption{\textbf{Example time series window from the synthetic data of the simulation study:} $x_{c,p}$ denotes $p$th patch in $c$th modality. Note how patches from different modalities within the same time window exhibit strong cross-modality correlations.} 
\label{simulation_data}
\end{figure}

\subsection{Clinical Relevance}

For digital health applications, this means the proposed MoCA approach more effectively exploits both within-modality temporal associations and cross-modality relationships in wearable data.
By leveraging these rich dependencies during pre-training, the models learn representations that are not only more robust to sensor dropout, but also better capture the complex physiological patterns that span multiple sensing modalities. 
This theoretical foundation provides a principled framework for designing efficient masking strategies that improve model robustness and clinical utility when learning from multi-modal wearable sensor data.



\section{Discussion}


\noindent \textbf{Summary}
\vspace{0.5mm}

In this work, we introduced our model Multi-modal Cross-masked Autoencoder (MoCA), a novel self-supervised learning framework for multi-modal wearable device data. MoCA integrates state-of-the-art transformer architectures with a principled cross-modality masking strategy to explicitly leverage both intra- and inter-modality correlation structures. The masking scheme in MoCA exploits the pronounced correlation structures in multi-modal health data, making it particularly well-suited for digital health applications and enabling the generation of informative latent representations for downstream tasks. Through extensive evaluation on real-world benchmark datasets, MoCA outperformed baselines in downstream tasks and achieved the best overall averages in transfer learning scenarios.

Beyond empirical performance, we established the first theoretical link between multi-modal MAE loss and kernelized canonical correlation analysis (KCCA) within a Reproducing Kernel Hilbert Space (RKHS) framework, offering a principled foundation for future designs of optimal masking strategies tailored to the correlation structures of wearable device data.

Our approach demonstrates MoCA's ability to learn transferable representations across diverse datasets by leveraging unlabeled data in pre-training and smaller labeled datasets in fine-tuning, indicating strong potential for applying masked self-supervised learning to wearable device data in both free-living environments and clinical studies. The framework's extensibility to emerging sensor modalities, including continuous heart rate, blood pressure, and glucose monitoring, positions it well for next-generation wearable technologies. As digital health evolves toward comprehensive multi-modal monitoring, principled approaches such as MoCA will be essential for translating raw sensor data into clinically actionable insights.



\vspace{3mm}
\noindent \textbf{Limitations and Caveats}

\vspace{0.5mm}
Our evaluation primarily focuses on accelerometer and physiological sensors. The generalizability to other sensor modalities (e.g., ECG, EMG, or environmental sensors) requires further validation, particularly for clinical-grade devices with different noise characteristics and sampling rates.

Interpretability remains a central challenge in deep learning approaches for digital health, especially for translating model outputs into health guidelines and personalized recommendations to promote healthy lifestyles. In this work, the use of attention maps provides a first step toward visualizing modality- and patch-level contributions to activity classification. Future work will refine these interpretability mechanisms, e.g., through clinically-informed attention constraints to enhance interpretability and facilitate clinical decision-making.

Another major challenge is the heterogeneity in free-living wearable device data, especially those from clinical studies. For example, heterogeneity can arise from differences in devices, body-worn locations, sampling protocols, and population characteristics (e.g., with or without specific medical conditions). Without careful treatment, such heterogeneity can introduce bias and limit model robustness. These challenges highlight the importance of pre-training on larger, more diverse datasets such as National Health and Nutrition Examination Survey (NHANES) (\url{https://wwwn.cdc.gov/nchs/nhanes/default.aspx}) to improve transferability and fairness.

\vspace{3mm}
\noindent \textbf{Future Directions}

\vspace{0.5mm}
The theoretical framework developed in this work provides a foundation for more sophisticated masking schemes. In our future research, we will explore correlation-adaptive masking that dynamically adjusts masking patterns based on the observed inter-modal correlation structure. This approach could be particularly valuable for handling non-stationary environments where the relationship between sensor modalities evolves over time. 

Another key future direction will be fine-tuning on clinical study data, which are often collected in free-living environments and thus tend to be noisier, more heterogeneous, and prone to device- and population-specific biases. Addressing these challenges will require heterogeneity-aware modeling strategies that explicitly account for variability across devices, body-worn locations, populations, and measurement protocols. Such developments will be critical for reducing potential bias, improving model robustness, and ensuring fair and reliable application of self-supervised learning methods in real-world clinical applications.


\section{Data availability}
This project uses the Human Activity Recognition Using Smartphones Dataset (UCIHAR) \cite{ucihar}. As per their dataset license, Creative Commons Attribution 4.0 International (CC BY 4.0) license, part of the pre-processed dataset is shared at \url{https://github.com/HowonRyu/MoCA} for demonstration purposes.

Benchmark datasets (WISDM, IMWSHA, OPPORTUNITY, ADL, PAMAP2, and RealWorld) used for transfer learning evaluation were released by \cite{harnet} and are available at \url{https://github.com/OxWearables/ssl-wearables}.

\section{Code availability}
The code is available at \url{https://github.com/HowonRyu/MoCA}.

\section{Funding}
This work was supported by NIH/NHLBI R01HL166802 (support to H. Ryu, Y. Chen, A. LaCroix, L. Natarajan, and J. Zou). The funder played no role in study design, data collection, analysis and interpretation of data, or the writing of this manuscript.

\section{Author contributions}
Howon Ryu and Yuliang Chen contributed equally as co-first authors to methodology development, formal analysis, and drafting. Yacun Wang maintained the computing infrastructure and assisted with GPU-based cluster computing. 

Yu Wang (\texttt{feather1014@gmail.com}) and Jingjing Zou (\texttt{j2zou@ucsd.edu}) serve as co-corresponding authors; they conceived the study, supervised the project, and led manuscript drafting and revisions. Jingjing Zou is responsible for writing the theoretical analysis. Andrea LaCroix, Chongzhi Di, and Loki Natarajan contributed to study design, data curation, and interpretation. All authors revised and approved the final manuscript.

\section{Competing interests}
The authors declare no competing interests.

\printbibliography

\clearpage

\appendix
\renewcommand{\thesection}{\Alph{section}}
\section*{Appendices}
\addcontentsline{toc}{section}{Appendix}

\section{Dataset}
\label{dataset}
Table \ref{tab:dataset_summary} contains a summary of all the datasets we used. For pre-training, we used UCI-HAR and segmented each time series using a 4-second sliding window. If a window contains two different activity labels and each activity is present for more than 1.2 seconds, we assign the label ``transition'' to that segment, resulting in a 7-class classification task. We do not apply any additional pre-processing such as normalization. The dataset was split into training and test sets with a 7:3 ratio. For transfer learning, we include WISDM, OPPORTUNITY, ADL, PAMAP2, and RealWorld, which are benchmark datasets processed and released by \cite{harnet}. To better match our pre-training setup, we re-segment their 10-second windows into 5-second windows and upsample the 30 Hz signals to 50 Hz. Note that when using MoCA for classification, the input does not need to match the length used during pre-training, because transformer models can naturally handle sequences of different lengths. In our case, we use a 5-second input to remain close to the distribution of the pretraining data and to allow a fair comparison with \cite{harnet}, which was originally developed for 5-second intervals that are not easily adapted to 4-second input. Additionally, we include IMWSHA, which contains both accelerometer and gyroscope like UCI-HAR.

\begin{table}[ht]
\centering
\caption{\textbf{Summary of datasets used for pre-training and evaluation.} 
The table lists the number of subjects, samples, activity classes, set usage, activity labels, sensor modalities, and references for each dataset.}
\resizebox{\textwidth}{!}{
\begin{tabular}{@{}lcccc>{\raggedright\arraybackslash}p{5.5cm}>{\raggedright\arraybackslash}p{4cm}l@{}}
\toprule
\textbf{Dataset} & \textbf{\#Subjects} & \textbf{\#Samples} & \textbf{\#Classes} & \textbf{Set-Type} & \textbf{Label Activities} & \textbf{Modalities} & \textbf{Reference} \\
\midrule
UCI-HAR & 30 & 1.8k & 7 & Pre-train / Eval & walking, walking upstairs, walking downstairs, sitting, standing, laying, transition & \makecell[l]{Accelerometer X,Y,Z\\Gyroscope X,Y,Z} & \cite{ucihar} \\
\midrule
IMWSHA & 10 & 0.3k & 11 & Eval & computer use, phone conversation, vacuuming, reading, watching TV, ironing, walking, exercising, cooking, drinking, brushing hair & \makecell[l]{Accelerometer X,Y,Z\\Gyroscope X,Y,Z} & \cite{imwsha1,imwsha2} \\
\midrule
WISDM & 46 & 56k & 6 & Eval & walking, jogging, walking upstairs, walking downstairs, sitting, standing & Accelerometer X,Y,Z & \cite{wisdm} \\
\midrule
RealWorld & 14 & 24k & 8 & Eval & climbing down, climbing up, jumping, lying, running, sitting, standing, walking & Accelerometer X,Y,Z & \cite{realworld} \\
\midrule
OPPORTUNITY & 4 & 7.8k & 4 & Eval & lying down, standing, sitting, walking & Accelerometer X,Y,Z & \cite{oppo} \\
\midrule
PAMAP2 & 8 & 5.8k & 8 & Eval & lying down, sitting, standing, ironing, vacuum cleaning, walking, ascending stairs, descending stairs & Accelerometer X,Y,Z & \cite{pamap} \\
\midrule
ADL & 7 & 1.2k & 5 & Eval & climb stairs, drink glass, get up from bed, pour water, walk & Accelerometer X,Y,Z & \cite{adl} \\
\bottomrule
\end{tabular}
}
\label{tab:dataset_summary}
\end{table}

\section{Implementation Details}
\label{implementation}
The hyperparameter configurations for pre-training, fine-tuning, and linear probing are detailed in Tables \ref{tab:pretrain_hyper}, \ref{tab:finetune_hyper}, and \ref{tab:lp_hyper}, respectively, following established practices and adapted from the DeiT \cite{codebase} codebase.

Table \ref{tab:pretrain_hyper} summarizes the hyperparameters used for pre-training using the UCI-HAR dataset. It specifies that AdamW serves as the optimizer, with an absolute learning rate of 5e-4 and a weight decay of 0.05. The momentum parameters are $\beta_1$=0.9 and $\beta_2$=0.95, and the batch size is 50. A cosine decay learning rate schedule is used, coupled with 50 warm-up epochs. The model is trained for 20 epochs if data augmentation is applied, or 4,000 epochs if no augmentation is applied. Table \ref{tab:finetune_hyper} and Table \ref{tab:lp_hyper} show the fine-tuning and linear probing settings for different datasets.
\begin{table*}[ht]
\centering
\begin{tabular}{@{}p{0.38\textwidth}@{\hspace{0.02\textwidth}}p{0.60\textwidth}@{}}

\begin{minipage}[t]{\linewidth}
\captionof{table}{\textbf{Pre-training settings}}
\label{tab:pretrain_hyper}
\resizebox{\textwidth}{!}{%
\begin{tabular}{l|l}
\toprule
\textbf{Configuration} & \textbf{Value} \\
\midrule
Optimizer & AdamW \\
Absolute Learning Rate & 5e-4 \\
Weight Decay & 5e-2 \\
Optimizer Momentum & $\beta_1=0.9, \beta_2=0.95$ \\
Batch Size & 50 \\
Learning Rate Schedule & Cosine Decay \\
Warm-up Epochs & 50 \\
Training Epochs & \begin{tabular}[t]{@{}l@{}}20 (w/ aug) \\ 4,000 (w/o aug)\end{tabular} \\
\bottomrule
\end{tabular}
}
\end{minipage}
&

\begin{minipage}[t]{\linewidth}
\captionof{table}{\textbf{Fine-tuning settings}}
\label{tab:finetune_hyper}
\resizebox{\textwidth}{!}{%
\begin{tabular}{l|c|c|c|c|c|c}
\toprule
\textbf{Configuration} & UCI-HAR & WISDM & IMWSHA & RealWorld & OPPORTUNITY & PAMAP2 \\
\midrule
Optimizer & AdamW & AdamW & AdamW & AdamW & AdamW & AdamW \\
Learning Rate & 1e-3 & 1e-3 & 2.5e-4 & 2.5e-4 & 2.5e-4 & 2.5e-4 \\
Weight Decay & 5e-2 & 5e-2 & 5e-2 & 5e-2 & 1e-1 & 5e-2 \\
Optimizer Momentum & \multicolumn{6}{c}{$\beta_1=0.9, \beta_2=0.999$} \\
Batch Size & 50 & 256 & 64 & 64 & 64 & 64 \\
LR Schedule & Cosine Decay & Cosine Decay & Cosine Decay & Cosine Decay & Cosine Decay & Cosine Decay \\
Warm-up Epochs & 5 & 10 & 10 & 10 & 10 & 10 \\
Training Epochs & 50 & 50 & 50 & 50 & 50 & 50 \\
\bottomrule
\end{tabular}
}

\vspace{1em}

\captionof{table}{\textbf{Linear probing settings}}
\label{tab:lp_hyper}
\resizebox{\textwidth}{!}{%
\begin{tabular}{l|c|c|c|c|c|c}
\toprule
\textbf{Configuration} & UCI-HAR & WISDM & IMWSHA & RealWorld & OPPORTUNITY & PAMAP2 \\
\midrule
Optimizer & AdamW & AdamW & AdamW & AdamW & AdamW & AdamW \\
Learning Rate & 1e-3 & 1e-2 & 1e-3 & 2.5e-3 & 2.5e-3 & 2.5e-3 \\
Weight Decay & 0 & 1e-4 & 1e-4 & 1e-4 & 5e-2 & 1e-4 \\
Optimizer Momentum & \multicolumn{6}{c}{$\beta_1=0.9, \beta_2=0.999$} \\
Batch Size & 50 & 256 & 28 & 64 & 64 & 64 \\
LR Schedule & Cosine Decay & Cosine Decay & Cosine Decay & Cosine Decay & Cosine Decay & Cosine Decay \\
Warm-up Epochs & 10 & 10 & 10 & 10 & 10 & 10 \\
Training Epochs & 50 & 50 & 50 & 50 & 50 & 50 \\
\bottomrule
\end{tabular}
}
\end{minipage}

\end{tabular}
\end{table*}

\section{Additional Metrics}
\label{additional_metrics}

Several of the evaluation datasets are class-imbalanced (e.g., OPPORTUNITY and ADL have activities that occur far more frequently than others), so top-1 accuracy can be dominated by the majority classes and overstate performance on rare activities. To complement the top-1 accuracy reported in Table~\ref{tab:kfold}, we report two metrics that weight all classes equally: \emph{macro-F1}, the unweighted mean of per-class F1 scores (Table \ref{tab:macro_f1}), and \emph{balanced accuracy}, the mean of per-class recall (Table \ref{tab:balanced_acc}). Both are reported as mean $\pm$ standard deviation across the subject-wise cross-validation folds, under linear probing (LP), where the backbone is frozen and only a linear head is trained, and fine-tuning (FT), where the full network is updated. Gray shading indicates the best performance in each dataset within each LP/FT block.

\begin{table*}[ht]
\centering
\caption{\textbf{Macro-F1 (mean $\pm$ standard deviation) across six activity recognition datasets under linear probing (LP) and fine-tuning (FT).} Macro-F1 is the unweighted mean of per-class F1 scores. Gray shading indicates the best performance in each dataset within each block.}
\label{tab:macro_f1}
\resizebox{\textwidth}{!}{
\begin{tabular}{@{}llcccccc@{}}
\toprule
Model & Setting & WISDM & IMWSHA & OPPORTUNITY & ADL & PAMAP2 & RealWorld \\
\midrule
MAE Synchronized Masking (w/o aug.) & LP & 59.42 $\pm$ 5.15 & 44.76 $\pm$ 7.62 & 33.18 $\pm$ 11.77 & 54.05 $\pm$ 5.48 & 62.59 $\pm$ 3.36 & 65.74 $\pm$ 4.86 \\
MAE Synchronized Masking (w/ aug.) & LP & 66.70 $\pm$ 3.65 & 57.16 $\pm$ 6.06 & 35.53 $\pm$ 6.98 & 58.95 $\pm$ 1.14 & 66.96 $\pm$ 3.33 & 67.66 $\pm$ 6.21 \\
Yuan et al. \cite{harnet} (pre-trained on UCI-HAR) & LP & 58.10 $\pm$ 4.57 & 29.34 $\pm$ 6.06 & 30.58 $\pm$ 5.14 & 25.70 $\pm$ 5.34 & 48.24 $\pm$ 4.96 & 65.34 $\pm$ 3.88 \\
MoCA (w/o aug.) & LP & 65.87 $\pm$ 4.38 & 59.81 $\pm$ 3.73 & 41.09 $\pm$ 9.81 & 54.46 $\pm$ 2.79 & 65.23 $\pm$ 2.10 & 71.32 $\pm$ 3.36 \\
\textbf{MoCA (w/ aug.)} & LP & \cellcolor{gray!30} \textbf{69.30 $\pm$ 4.22} & \cellcolor{gray!30} \textbf{67.00 $\pm$ 10.36} & \cellcolor{gray!30} \textbf{41.90 $\pm$ 9.44} & \cellcolor{gray!30} \textbf{65.44 $\pm$ 3.83} & \cellcolor{gray!30} \textbf{75.00 $\pm$ 4.73} & \cellcolor{gray!30} \textbf{72.89 $\pm$ 3.66} \\
\cmidrule{1-8}
MAE Synchronized Masking (w/o aug.) & FT & 65.97 $\pm$ 4.12 & 64.07 $\pm$ 7.39 & 39.08 $\pm$ 7.75 & 62.69 $\pm$ 4.80 & 71.09 $\pm$ 2.22 & 75.52 $\pm$ 4.84 \\
MAE Synchronized Masking (w/ aug.) & FT & 71.09 $\pm$ 3.99 & 73.07 $\pm$ 10.61 & 42.88 $\pm$ 3.10 & 62.62 $\pm$ 2.30 & 72.43 $\pm$ 2.29 & 74.66 $\pm$ 3.07 \\
Yuan et al. \cite{harnet} (pre-trained on UCI-HAR) & FT & \cellcolor{gray!30} \textbf{76.20 $\pm$ 3.85} & 75.82 $\pm$ 7.57 & 40.82 $\pm$ 12.14 & \cellcolor{gray!30} \textbf{85.16 $\pm$ 2.26} & 74.87 $\pm$ 3.37 & \cellcolor{gray!30} \textbf{80.65 $\pm$ 5.85} \\
MoCA (w/o aug.) & FT & 70.91 $\pm$ 4.20 & 71.73 $\pm$ 9.37 & \cellcolor{gray!30} \textbf{49.27 $\pm$ 7.77} & 69.22 $\pm$ 4.52 & 70.53 $\pm$ 3.03 & 75.09 $\pm$ 3.52 \\
\textbf{MoCA (w/ aug.)} & FT & 74.54 $\pm$ 4.02 & \cellcolor{gray!30} \textbf{76.03 $\pm$ 7.35} & 44.83 $\pm$ 5.59 & 75.96 $\pm$ 3.09 & \cellcolor{gray!30} \textbf{78.53 $\pm$ 4.23} & 80.06 $\pm$ 2.79 \\
\bottomrule
\end{tabular}
}
\end{table*}

\begin{table*}[ht]
\centering
\caption{\textbf{Balanced accuracy (mean $\pm$ standard deviation) across six activity recognition datasets under linear probing (LP) and fine-tuning (FT).} Balanced accuracy is the mean of per-class recall. Gray shading indicates the best performance in each dataset within each block.}
\label{tab:balanced_acc}
\resizebox{\textwidth}{!}{
\begin{tabular}{@{}llcccccc@{}}
\toprule
Model & Setting & WISDM & IMWSHA & OPPORTUNITY & ADL & PAMAP2 & RealWorld \\
\midrule
MAE Synchronized Masking (w/o aug.) & LP & 60.10 $\pm$ 5.66 & 48.50 $\pm$ 8.22 & 38.22 $\pm$ 9.00 & 56.47 $\pm$ 3.17 & 62.39 $\pm$ 3.84 & 65.78 $\pm$ 5.18 \\
MAE Synchronized Masking (w/ aug.) & LP & 67.26 $\pm$ 3.97 & 61.21 $\pm$ 6.39 & 40.33 $\pm$ 7.34 & 59.42 $\pm$ 0.65 & 67.79 $\pm$ 2.65 & 68.62 $\pm$ 5.81 \\
Yuan et al. \cite{harnet} (pre-trained on UCI-HAR) & LP & 58.98 $\pm$ 4.86 & 35.26 $\pm$ 4.48 & 34.64 $\pm$ 5.17 & 30.35 $\pm$ 5.83 & 51.29 $\pm$ 3.76 & 64.14 $\pm$ 3.84 \\
MoCA (w/o aug.) & LP & 66.31 $\pm$ 4.68 & 62.64 $\pm$ 3.62 & 46.46 $\pm$ 7.22 & 55.54 $\pm$ 3.19 & 65.41 $\pm$ 2.09 & 71.46 $\pm$ 3.62 \\
\textbf{MoCA (w/ aug.)} & LP & \cellcolor{gray!30} \textbf{69.51 $\pm$ 4.69} & \cellcolor{gray!30} \textbf{70.21 $\pm$ 10.89} & \cellcolor{gray!30} \textbf{47.81 $\pm$ 4.52} & \cellcolor{gray!30} \textbf{66.02 $\pm$ 3.88} & \cellcolor{gray!30} \textbf{74.99 $\pm$ 4.16} & \cellcolor{gray!30} \textbf{72.64 $\pm$ 4.25} \\
\cmidrule{1-8}
MAE Synchronized Masking (w/o aug.) & FT & 66.36 $\pm$ 4.45 & 66.13 $\pm$ 7.29 & 40.57 $\pm$ 5.57 & 63.88 $\pm$ 4.69 & 71.58 $\pm$ 2.02 & 75.47 $\pm$ 4.87 \\
MAE Synchronized Masking (w/ aug.) & FT & 71.12 $\pm$ 4.41 & 74.45 $\pm$ 10.80 & 45.48 $\pm$ 1.63 & 64.48 $\pm$ 1.49 & 73.06 $\pm$ 1.62 & 74.49 $\pm$ 3.07 \\
Yuan et al. \cite{harnet} (pre-trained on UCI-HAR) & FT & \cellcolor{gray!30} \textbf{76.35 $\pm$ 4.18} & 76.41 $\pm$ 8.40 & 43.55 $\pm$ 7.48 & \cellcolor{gray!30} \textbf{85.13 $\pm$ 2.18} & 75.38 $\pm$ 2.82 & \cellcolor{gray!30} \textbf{80.46 $\pm$ 6.02} \\
MoCA (w/o aug.) & FT & 71.02 $\pm$ 4.51 & 72.55 $\pm$ 9.36 & \cellcolor{gray!30} \textbf{49.86 $\pm$ 5.23} & 69.05 $\pm$ 4.28 & 70.68 $\pm$ 3.07 & 75.47 $\pm$ 3.85 \\
\textbf{MoCA (w/ aug.)} & FT & 74.53 $\pm$ 4.32 & \cellcolor{gray!30} \textbf{77.22 $\pm$ 8.73} & 46.65 $\pm$ 3.13 & 74.78 $\pm$ 3.35 & \cellcolor{gray!30} \textbf{78.34 $\pm$ 4.58} & 80.20 $\pm$ 2.75 \\
\bottomrule
\end{tabular}
}
\end{table*}

The two metrics tell a consistent story. Under \textbf{linear probing}, where the quality of the frozen representation is measured directly, MoCA with augmentation attains the best macro-F1 and balanced accuracy on all six datasets, confirming that the improvement is not an artifact of majority-class dominance but reflects genuinely better per-class discrimination. The margin over the synchronized-masking MAE is largest on the more imbalanced or fine-grained datasets (e.g., $+9.8$ macro-F1 on IMWSHA and $+8.0$ on PAMAP2), and the fully self-supervised baseline of Yuan et al. trails substantially in the frozen setting, indicating that MoCA's cross-modal masking learns representations that transfer without any backbone adaptation. Augmentation is consistently beneficial for MoCA at the LP stage, adding several points of macro-F1 across the board.

Under \textbf{fine-tuning}, where the entire backbone is adapted to each downstream task, the gap between methods narrows, as expected once the representation can be reshaped by the target labels. MoCA with augmentation remains the strongest self-supervised approach and posts the best scores on IMWSHA and PAMAP2, while MoCA without augmentation is best on OPPORTUNITY, the most imbalanced dataset, where equal-weighted metrics reward its more uniform per-class recall. Yuan et al. is competitive after fine-tuning and leads on WISDM, ADL, and RealWorld; ADL in particular is a small dataset where their supervised pre-training is well matched. Overall, both class-balanced metrics corroborate the top-1 accuracy results in the main text: MoCA's advantage is clearest where it matters most, in the linear-probing regime that isolates representation quality and on the imbalanced datasets where per-class performance is easily masked by aggregate accuracy.

\section{Supplementary Visualization}
\label{supp_vis}
\begin{figure}[t]
\centering
\caption{\textbf{Confusion matrices for UCI-HAR classification task using fine-tuned models.} Rows denote true labels and columns denote predicted labels.}
\begin{minipage}{0.32\textwidth}
    \centering
    \includegraphics[width=\textwidth]{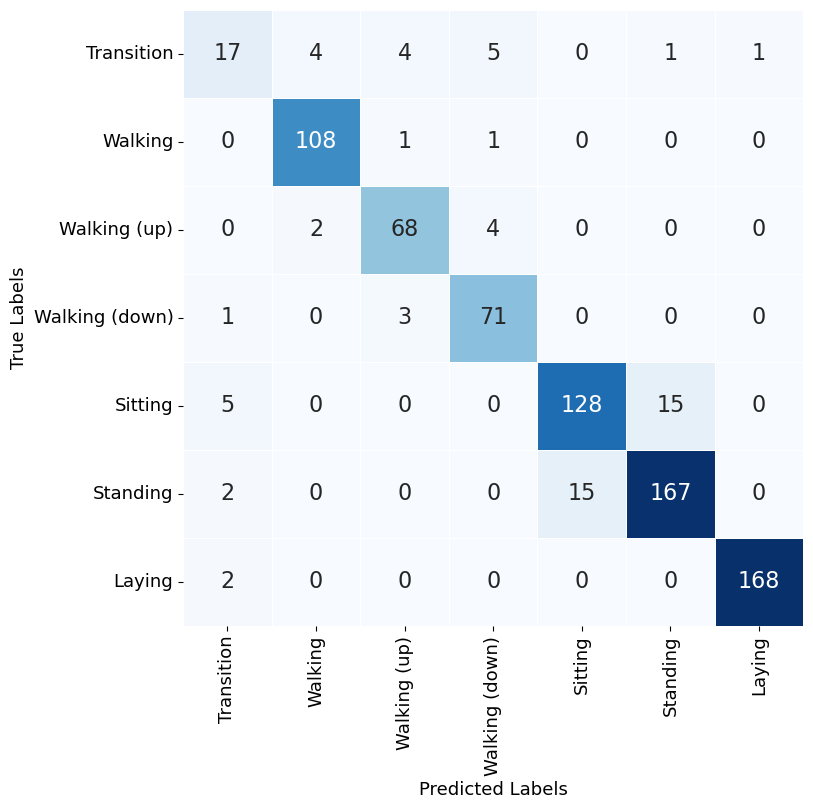}
    \caption*{(a) Synchronized masking}
\end{minipage}
\hfill
\begin{minipage}{0.32\textwidth}
    \centering
    \includegraphics[width=\textwidth]{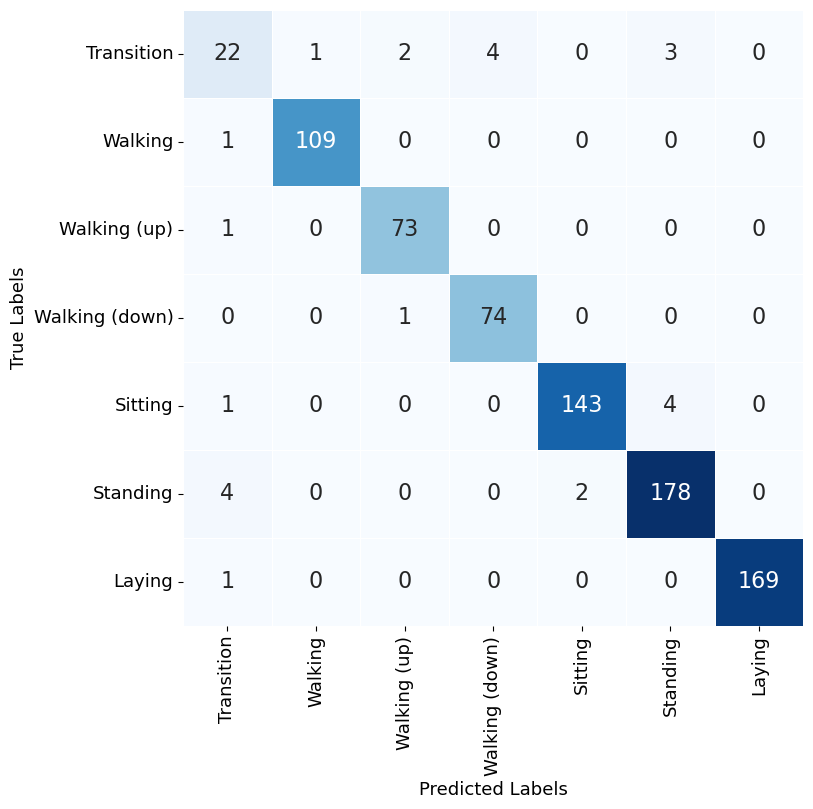}
    \caption*{(b) MoCA w/ augmentation (Ours)}
\end{minipage}
\hfill
\begin{minipage}{0.32\textwidth}
    \centering
    \includegraphics[width=\textwidth]{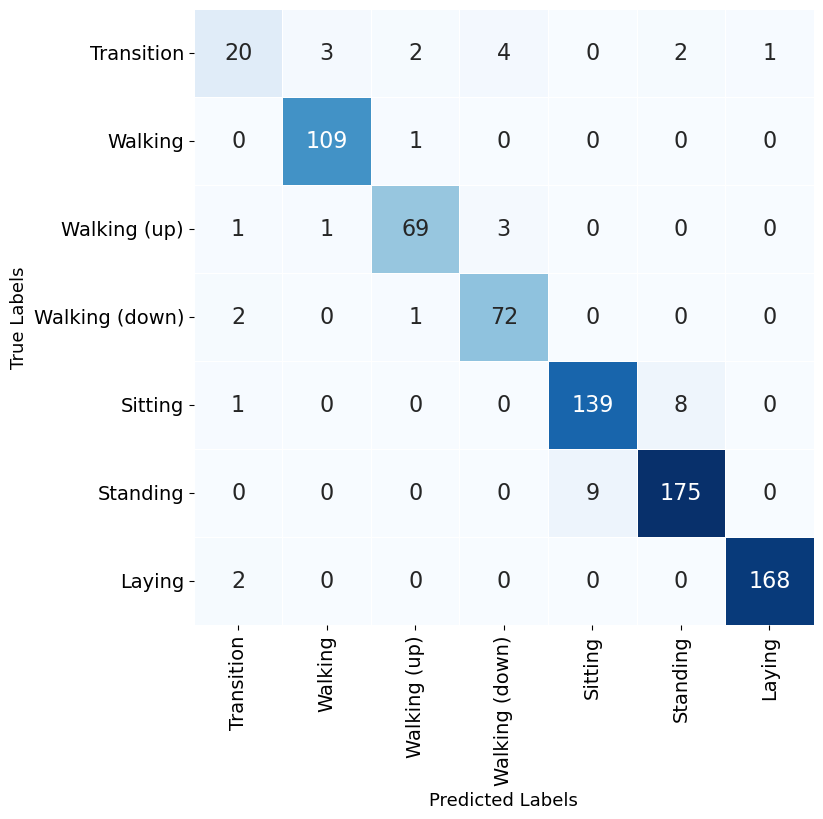}
    \caption*{(c) MoCA w/o augmentation (Ours)}
\end{minipage}
\label{fig:apdx_cm}
\end{figure}

We further visualize the classification results of MoCA using confusion matrices, which provide detailed insights into the models' performances by displaying counts of correct and incorrect classifications across different activity classes. As shown in Figure \ref{fig:apdx_cm}, MoCA achieves better classification accuracy than Synchronized masking MAE, especially in discerning activities with similar movement content, such as sitting and standing. 
MoCA with augmentation achieves a top-1 classification accuracy of 96.8\%, compared to 91.7\% for synchronized masking and 94.8\% for MoCA without augmentation, demonstrating the robustness of the learned representations using MoCA.

\section{Theoretical Analysis} \label{sect: theory}

In this section, we outline a theoretical analysis of the proposed method and establish a connection between the MAE reconstruction loss during pre-training and kernel canonical correlation analysis (KCCA) within a Reproducing Kernel Hilbert Space (RKHS) framework.

\subsection{Notation}
Let $\{x_i: i = 1,\dots, n\}$ denote the set of $n$ generalized “images” from a dataset. In our setting, an image is a 4-second window of multi-modal time-series data recorded by one or multiple concurrently worn wearable devices. Note that in our setting, the individual axes of a sensor are treated as distinct modalities.
Suppose there are $C$ modalities (rows), each split into $P$ contiguous, non-overlapping patches of fixed length $L_p$ (columns).  Hence every image contains $C\times P$ patches.
We write the $(c,p)$-th patch as
$x_{i,(c,p)}\in\mathbb R^{L_p}$, where $c=1,\dots, C$ and $\;p=1,\dots, P$.

For the $i$th image, 
let $m_i\in\{0,1\}^{C\times P}$
be the binary masking matrix, with entries $m_{i,(c,p)}$. A value of $1$ indicates that all elements in the corresponding patch are masked and $0$ indicates elements unmasked. The masked and unmasked “views” of $x_i$ are then defined as collections of all masked and unmasked patches in the image, respectively:
$$
x_i^{M}=x_i\odot m_i,\qquad   
x_i^{U}=x_i\odot(\mathbf 1-m_i),
$$
where $\odot$ denotes patch-wise (Hadamard) multiplication and $\mathbf 1$ is the all-ones matrix of the same shape as $m_i$.

Fix a mask ratio $\rho\in(0,1)$. For all $i$, realizations of $m_i$ must satisfy the constraint that
$$
\sum_{c=1}^C\sum_{p=1}^P m_{i,(c,p)}=\lfloor \rho\, CP\rfloor.
$$
Let
$$
\mathcal M_\rho=\Bigl\{m\in\{0,1\}^{C\times P}:\textstyle\sum_{c,p}m_{(c,p)}=\lfloor\rho\, CP\rfloor\Bigr\}
$$
be the set of admissible masks under mask ratio $\rho$. Conditional on $x_i$, the sampling scheme specifies a probability measure $P(\cdot\mid x_i)$ supported on $\mathcal M_\rho$; drawing $m_i\sim P(\cdot\mid x_i)$ independently for each $i$ determines which patches form the masked view $x_i^{M}$ and which remain in the unmasked view $x_i^{U}$.
Additional structural constraints can be imposed by further restricting the support of $P$.
For example, in a synchronized masking scheme, all modalities sharing the same time interval receive the same mask indicator, so an entire temporal slice is either fully masked or fully unmasked across modalities.

\subsection{Bound for MAE Reconstruction Loss}

The Masked Autoencoder (MAE) model consists of an encoder $f$ and a decoder $g$. 
Given a masking matrix $m_i$ and its induced partition $(x_i^{U},x_i^{M})$, the encoder $f$ maps the unmasked view to a latent feature $z_i^{U}$, which the decoder $g$ then maps to a reconstructed masked view:
$$
z_i^{U}=f\!\left(x_i^{U}\right),
\qquad   
\hat x_i^{M}=g\!\left(z_i^{U}\right)=(g\!\circ \! f)\!\left(x_i^{U}\right),
$$
where we define \(h := g \!\circ\! f\) for convenience.

\begin{assumption}[normalization] \label{assump 1}
All views and reconstructions are $\ell_2$-normalized \cite{mae, zhang2022mask}:
$$
\|x_i^{U}\|_2=\|x_i^{M}\|_2=\|\hat x_i^{M}\|_2, \quad\text{for all }i.
$$
\end{assumption}

The loss of the MAE is the expected mean-squared reconstruction error on the masked views:
\begin{equation} \label{eq: MAE loss}
L_{\text{MAE}} =\mathbb E_{x_i}
  \mathbb E_{m_i\sim P(\cdot\mid x_i)}
  \Bigl[
      \bigl\|
        \bigl(g\!\circ \!f\bigr)(x_i^{U})-x_i^{M}
      \bigr\|_2^{2}
  \Bigr],
\end{equation}
where the inner expectation is taken with respect to the masking distribution $P$ introduced earlier.  
In practice, \eqref{eq: MAE loss} is approximated by a Monte Carlo average over mini batches of samples and independently drawn masks. Hence, we write the empirical distribution version of the MAE loss  function as:
\begin{equation} \label{eq: MAE loss emp}
\widehat L_{\text{MAE}}
=\frac{1}{n}\sum_{i=1}^{n}
  \bigl\|
        \bigl(g\!\circ \! f\bigr)\!\bigl(x_i^{U}\bigr)
        -x_i^{M}
  \bigr\|_{2}^{2}.
\end{equation}

The following assumption is from \cite{zhang2022mask}:

\begin{assumption}[]
\label{Zhang3.1}
For any non-degenerate decoder $g$, there exists a pseudo-inverse encoder $f_g$ such that the pseudo autoencoder $h_g := g \circ f_g$ satisfies  $\mathbb{E}_x \|h_g(x) - x \|_2^2 \le \epsilon$, where $x$ denotes either the unmasked view $x^U$ or the masked view $x^M$ and $\epsilon>0$ is a small constant.
\end{assumption}

\begin{remark}
Assumption~\ref{Zhang3.1} requires that, for any fixed decoder \(g\), there exists a 
\emph{pseudo-inverse} encoder \(f_g\) that produces latent features which \(g\) can map back to a close approximation of the original input \(x\) (masked or unmasked view).  
The reconstruction need not be exact; it only needs to be within a small mean-squared error \(\epsilon\).  
\end{remark}

It was shown in \cite{zhang2022mask} that under Assumptions \ref{assump 1} and \ref{Zhang3.1}, the MAE loss has a lower bound:
\begin{equation} \label{eq: L align}
L_{\mathrm{MAE}}\;\ge\;-
      \mathbb E_{x}\,
      \mathbb E_{m\sim P(\cdot\mid x)}
      \bigl\langle
           h(x^{U}),
           h_{g}(x^{M})
      \bigr\rangle -\varepsilon
      +\text{const},
\end{equation}
where $\langle\cdot,\cdot\rangle$ denotes the Euclidean inner product.  The constant term depends only on the data distribution and the masking scheme, not on the learnable encoder $f$ and decoder $g$. 
Therefore, minimizing this bound is equivalent to maximizing the alignment term
$\mathbb E_{x} \mathbb E_{m\sim P(\cdot\mid x)}
\langle h(x^{U}), h_{g}(x^{M}) \rangle$.

Given $n$ samples and independently drawn masks $\{m_i\}$, the alignment term is approximated by the empirical distribution average
\begin{equation}\label{eq:empirical bound}
\frac{1}{n} \sum_{i=1}^n \langle h(x_i^U), h_g(x_i^M) \rangle.
\end{equation}

\begin{remark}
This lower bound is in fact tight under the $\ell_2$-normalization assumption. 
Assume decoder outputs are $\ell_2$-normalized to a fixed $c>0$ so that
$\|h(x^U)\|_2=\|h_g(x^M)\|_2=c$, and let $A:=\mathbb{E}\!\left[h(x^U)^\top h_g(x^M)\right]$.
If Assumption~\ref{Zhang3.1} holds with $\mathbb{E}\|h_g(x^M)-x^M\|_2^2\le\epsilon$, then
\begin{equation}\label{eq:equiv-band}
L_{\mathrm{MAE}}
\;=\; 2c^2 - 2A \;+\; \xi, 
\qquad
|\xi| \;\le\; \epsilon \;+\; 2\sqrt{\epsilon\,\bigl(2c^2-2A\bigr)}.
\end{equation}
In particular, when $\epsilon=0$ (perfect pseudo-inverse), we have the exact identity
$L_{\mathrm{MAE}}=2c^2-2A$; for small $\epsilon$, $L_{\mathrm{MAE}}$ and the alignment
$A$ are equivalent up to $O(\sqrt{\epsilon})$.

\begin{proof}[Sketch of Proof]
Write
\[
\|h(x^U)-x^M\|^2
=\|h-h_g+h_g-x^M\|^2
=\|h-h_g\|^2+\|h_g-x^M\|^2+2(h-h_g)^\top(h_g-x^M).
\]
Take expectations, use $\mathbb{E}\|h_g-x^M\|^2\le\epsilon$ and Cauchy–Schwarz:
\[
\Big|\mathbb{E}\big[(h-h_g)^\top(h_g-x^M)\big]\Big|
\;\le\; \sqrt{\mathbb{E}\|h-h_g\|^2}\,\sqrt{\mathbb{E}\|h_g-x^M\|^2}
\;\le\; \sqrt{\epsilon\,\mathbb{E}\|h-h_g\|^2}.
\]
Under constant-norm outputs, 
\(\mathbb{E}\|h-h_g\|^2
=\mathbb{E}(\|h\|^2+\|h_g\|^2-2h^\top h_g)=2c^2-2A\).
Thus
\[
L_{\mathrm{MAE}}
=\bigl(2c^2-2A\bigr)\;+\;\mathbb{E}\|h_g-x^M\|^2
\;+\;2\,\mathbb{E}\big[(h-h_g)^\top(h_g-x^M)\big],
\]
which yields \eqref{eq:equiv-band} after bounding the last two terms by
\(\epsilon+2\sqrt{\epsilon\,(2c^2-2A)}\).
\end{proof}

\end{remark}

We next adopt an RKHS representation of the encoder–decoder mappings to express the maximization of this empirical alignment term as an equivalent kernel canonical correlation analysis (KCCA) objective.

\subsection{Reproducing Kernel Hilbert Space Representations}
Let $k_u$ be a positive-definite kernel that measures the similarity between any two unmasked views in $\{x^U_i\}$. Similarly, let $k_m$ be a positive-definite kernel that measures the similarity between any two masked views in $\{x^M_i\}$. 
Note that here both $k_u$ and $k_m$ are defined in the generalized case of kernels for vector-valued functions. That is, $k_u$ and $k_m$ are matrix-valued (operator-valued) positive-definite kernels: for any pair of inputs they return a $d\times d$ positive-definite matrix, where $d \ge 1$ is the dimensionality of the vectors $h(x_i^{U})$ and $h_g(x_i^{M})$.  

These kernels induce vector-valued Reproducing Kernel Hilbert Spaces (RKHS) $\mathcal{H}_U$ and $\mathcal{H}_M$, whose elements are functions defined on the unmasked and masked views, respectively. In certain settings one may adopt a shared similarity measure by taking $k_u=k_m$. For generality we allow here the two views to be equipped with distinct kernels.

\begin{remark}
When each view is composed of multiple patches, the view-level kernels \(k_u\) and \(k_m\) can be constructed from patch-level kernels via a mean embedding.  
Let \(\mathcal{P}(x_i^{U})\) denote the set of patches in the unmasked view \(x_i^{U}\), and let \(k_{\mathrm{patch}}\) be a positive-definite kernel between individual patches.  
The corresponding view-level kernel can be defined as
\[
k_u(x_i^{U}, x_j^{U})
= \frac{1}{|\mathcal{P}(x_i^{U})| \, |\mathcal{P}(x_j^{U})|}
  \sum_{p \in \mathcal{P}(x_i^{U})}
  \sum_{q \in \mathcal{P}(x_j^{U})}
    k_{\mathrm{patch}}(p, q),
\]
i.e., the average of all patch–patch kernel evaluations between the two views.  
An analogous definition applies to \(k_m\) for masked views.  
Equivalently, if \(\psi_{\mathrm{patch}}(\cdot)\) is the feature map associated with \(k_{\mathrm{patch}}\), the view-level feature embedding can be taken as the mean patch embedding
\[
\psi_{\mathrm{view}}(x_i^{U})
= \frac{1}{|\mathcal{P}(x_i^{U})|} 
  \sum_{p \in \mathcal{P}(x_i^{U})} \psi_{\mathrm{patch}}(p),
\]
so that \(k_u(x_i^{U}, x_j^{U}) = 
\langle \psi_{\mathrm{view}}(x_i^{U}),\, \psi_{\mathrm{view}}(x_j^{U}) \rangle\).  
This mean-embedding construction preserves positive-definiteness and naturally aggregates patch-level similarities into a view-level similarity.
\end{remark}

By the Representer Theorem \cite{scholkopfGeneralizedRepresenterTheorem2001, micchelliLearningVectorValuedFunctions2005b, alvarezKernelsVectorValuedFunctions2012}, given Assumption \ref{assump 1},
any $h \in \mathcal{H}_U$ and $h_g \in \mathcal{H}_M$ that maximize \eqref{eq:empirical bound} have the following representations:
\begin{equation} \label{eq: linear reps}
h(x_i^U) = \sum_{k=1}^n k_u(x_i^U, x_k^U) \alpha_k, \quad
h_g(x_i^M) = \sum_{k=1}^n k_m(x_i^M, x_k^M) \beta_k,
\end{equation}
with coefficient vectors $\alpha_k,\beta_k\in\mathbb R^{d}$ for $k=1,\dots,n$.
When $d=1$ (scalar outputs), these reduce to the familiar scalar-RKHS representations: weighted linear combinations of $\{k_u(\,\cdot\,,x_k^{U})\}$ and $\{k_m(\,\cdot\,,x_k^{M})\}$.

\begin{remark}
In this formulation we assume that the learnable mappings \(h = g \! \circ \! f\) and \(h_g = g \! \circ \! f_g\) can be well-approximated by elements of \(\mathcal{H}_U\) and \(\mathcal{H}_M\), respectively.  
In practice, this requires choosing kernels \(k_u\) and \(k_m\) whose associated RKHS are rich enough to capture the views' latent patterns as well as the similarities in semantics among the views, so that the encoder–decoder mappings are representable in the RKHS form \eqref{eq: linear reps}.
For kernel choice, see operator-valued kernels for vector-valued functions and multi-task learning \cite{micchelliLearningVectorValuedFunctions2005b,carmeliVectorValuedReproducing2010,alvarezKernelsVectorValuedFunctions2012}, and alignment/deformation-tolerant time series kernels such as \cite{cuturiKernelTimeSeries2007,andenDeepScatteringSpectrum2014,salviSignatureKernelSolution2021}.
\end{remark}

\subsection{Connection to Kernel Canonical Correlation Analysis}

Let \(K_U \in \mathbb{R}^{nd \times nd}\) denote the block Gram matrix whose \((i,j)\)-th block 
\(K_U^{(i,j)} \in \mathbb{R}^{d \times d}\) is given by \(k_u(x_i^{U}, x_j^{U})\).  
Similarly, let \(K_M \in \mathbb{R}^{nd \times nd}\) denote the block Gram matrix whose \((i,j)\)-th block is \(k_m(x_i^{M}, x_j^{M})\).  
When \(d=1\), \(K_U\) and \(K_M\) reduce to the standard \(n \times n\) scalar-valued Gram matrices.

With the representations \eqref{eq: linear reps}, the empirical alignment term \eqref{eq:empirical bound} can be expressed as
\begin{equation} \label{eq: CCA}
\frac{1}{n} \,\alpha^{\!\top} K_U K_M \beta,
\end{equation}
where \(\alpha = [\alpha_1, \dots, \alpha_n]^{\!\top}\) and \(\beta = [\beta_1, \dots, \beta_n]^{\!\top}\).  
Under Assumption~\ref{assump 1} and omitting the constant factor \(1/n\), the optimal solution to \eqref{eq: CCA} is obtained by solving
\begin{equation} \label{eq: kernel obj}
(\alpha^*, \beta^*) 
= \arg\max_{\alpha, \beta} 
\frac{\alpha^{\!\top} K_U K_M \beta}
{\sqrt{\alpha^{\!\top} K_U^{2} \alpha \; \beta^{\!\top} K_M^{2} \beta}},
\end{equation}
which is the \emph{uncentered kernel canonical correlation analysis} (KCCA) objective.  
When centering is appropriate (as in the example in the next section), one can replace each Gram matrix by its centered form
$\tilde K = H K H$ with $H = I - \tfrac{1}{n}\mathbf{1}\mathbf{1}^\top$, which corresponds to centering in feature space and yields the classical centered KCCA formulation. For more details, see, e.g., \cite{scholkopfGeneralizedRepresenterTheorem2001,fukumizu2007kcca,andrew2013deep}.

The KCCA problem is solved using Lagrange multipliers.
With the normalization constraints 
\(\alpha^\top K_U^{2} \alpha = 1\) and \(\beta^\top K_M^{2} \beta = 1\), 
the stationary equations from \eqref{eq: kernel obj} can be written as
\[
K_U K_M \beta = \rho \, K_U^{2} \alpha,
\qquad
K_M K_U \alpha = \rho \, K_M^{2} \beta.
\]
These can be combined into the block generalized eigenvalue problem
\begin{equation} \label{eq: kcca block}
\begin{bmatrix}
0 & K_U K_M \\
K_M K_U & 0
\end{bmatrix}
\begin{bmatrix}
\alpha \\ \beta
\end{bmatrix}
=
\rho
\begin{bmatrix}
K_U^{2} & 0 \\
0 & K_M^{2}
\end{bmatrix}
\begin{bmatrix}
\alpha \\ \beta
\end{bmatrix}.
\end{equation}
In practice, regularization is applied to avoid degeneracy by replacing 
\(K_U^{2}\) and \(K_M^{2}\) in \eqref{eq: kcca block} with
\[
K_U^{2} + \gamma_U K_U,
\qquad
K_M^{2} + \gamma_M K_M,
\]
where \(\gamma_U, \gamma_M > 0\) are small regularization parameters.  
The leading eigenvalue \(\rho\) gives the maximum objective, and the associated eigenvectors yield the optimal coefficients \(\alpha^*\) and \(\beta^*\).
For detailed derivations, see, e.g., \cite{hardoon2004canonical,fukumizu2007kcca,andrew2013deep, yang2019survey}.

\subsection{Influence of Masking Schemes}

A stochastic masking policy is specified by a probability measure
\(P(\,\cdot\,|x)\) on the admissible set \(\mathcal{M}_\rho\).
For each image \(x_i\), a mask \(m_i \sim P(\,\cdot\,|x_i)\) is drawn independently, and
\(\mathbf{m} = (m_1,\dots,m_n) \in \mathcal{M}_\rho^{\,n}\) collects the masks for a mini-batch or dataset.

Given \(\mathbf{m}\), define the block Gram matrices
\[
K_U(\mathbf{m}) = \bigl[k_u(x_i^{U}(\mathbf{m}),\,x_j^{U}(\mathbf{m}))\bigr]_{i,j},
\qquad
K_M(\mathbf{m}) = \bigl[k_m(x_i^{M}(\mathbf{m}),\,x_j^{M}(\mathbf{m}))\bigr]_{i,j}.
\]
Let
\[
\Phi(\mathbf{m}) :=
\max_{\alpha,\beta} \frac{\alpha^{\!\top}K_U(\mathbf{m})\,K_M(\mathbf{m})\,\beta}
{\sqrt{\alpha^{\!\top}K_U(\mathbf{m})^2\alpha}\;
 \sqrt{\beta^{\!\top}K_M(\mathbf{m})^2\beta}}
\]
be the optimal value of \eqref{eq: kernel obj} for this mask realization.
The population objective \eqref{eq: CCA} is \(\mathbb{E}_{\mathbf{m}}[\Phi(\mathbf{m})]\).
Once \(K_U(\mathbf{m})\) and \(K_M(\mathbf{m})\) are known or estimated from a pilot sample, choosing a masking scheme amounts to selecting \(P\) to put larger mass on masks with larger \(\Phi(\mathbf{m})\),
subject to the mask ratio and any structural constraints.

\paragraph{Linear scalar-kernel special case.}
With linear kernels,
\[
k_u(x_i^U, x_j^U) = \Phi_u(x_i^U)^\top \Phi_u(x_j^U),
\quad
k_m(x_i^M, x_j^M) = \Phi_m(x_i^M)^\top \Phi_m(x_j^M),
\]
where $\Phi_u$ and $\Phi_m$ map views to feature vectors representing their semantics.
It is recommended to choose \(\Phi_u,\Phi_m\) to be (approximately) alignment-invariant, so that views with similar semantics but different temporal indices (e.g., phase shifts/mis-alignment) still yield high kernel similarity.

In this case the sample cross-covariance and within-view covariance matrices are
$\Sigma_{U\!U} := \sum_{i=1}^n \Phi_u(x_i^U)\,\Phi_u(x_i^U)^\top /n$,
$\Sigma_{M\!M} := \sum_{i=1}^n \Phi_m(x_i^M)\,\Phi_m(x_i^M)^\top /n$
and 
$\Sigma_{U\!M} := \sum_{i=1}^n \Phi_u(x_i^U)\,\Phi_m(x_i^M)^\top /n$.
Then \eqref{eq: kernel obj} reduces to the CCA problem (deep CCA if $\Phi$ is a deep network model):
\begin{equation} \label{eq: classical CCA}
\max_{w_u, w_m}
\frac{w_u^\top \Sigma_{U\!M} w_m}
{\sqrt{(w_u^\top \Sigma_{U\!U} w_u) (w_m^\top \Sigma_{M\!M} w_m)}},
\end{equation}
whose optimum is the top singular value $\sigma_1$ of
\(\Gamma = \Sigma_{U\!U}^{-1/2} \Sigma_{U\!M} \Sigma_{M\!M}^{-1/2}\) \cite{sanchez1979multivariate, andrew2013deep}.
Thus, the optimal masking distribution concentrates on allocations yielding larger \(\sigma_1(\Gamma)\).
$\Gamma$ can be viewed as the biadjacency matrix of a weighted bipartite graph with unmasked and masked views as the two parts. Its top singular value $\sigma_1(\Gamma)$ is the spectral norm and quantifies the strongest global linear association across the two sides. A large $\sigma_1$ indicates strong cross-view structure \cite{PhysRevE}.

\subsection{Demonstration with Synthetic Data}

To validate the above heuristic, we compare MoCA against the synchronized masking scheme through a simulation with synthetic data.
Following the augmentation Algorithm \ref{alg:augment_time_series}, we generate synthetic data by sampling two samples $\bar{x}_1$ and $\bar{x}_2$ from the training set, and then randomly extract a consecutive chunk of time series from $\bar{x}_2$ to replace the corresponding time series in $\bar{x}_1$. The resulting synthetic time series represents a time window in which the participant transitions, e.g., from walking to sitting.
We repeated this process 1,000 times to obtain our simulation data. An example image is shown in Figure \ref{simulation_data}.

MoCA and synchronized masking schemes yield different partitions of the images into unmasked and masked views. Under each masking scheme, we map the unmasked and masked views to feature vectors with an encoder then reduce to 50 dimensions with PCA (scores are centered by construction).
Let $Z_U^{(s)},Z_M^{(s)}\in\mathbb{R}^{n\times 50}$ be the PC-score matrices. Define for each partition
$$
\Sigma_{U\!U}^{(s)}=\tfrac1n (Z_U^{(s)})^\top Z_U^{(s)},\quad
\Sigma_{M\!M}^{(s)}=\tfrac1n (Z_M^{(s)})^\top Z_M^{(s)},\quad
\Sigma_{U\!M}^{(s)}=\tfrac1n (Z_U^{(s)})^\top Z_M^{(s)}.
$$
We then compute the matrix $\Gamma$ and calculate the top singular value $\sigma_1$. The resulting top singular values $\sigma_1^{\text{MoCA}} = 0.925$ while $\sigma_1^{\text{Synchronize}} = 0.835$.

The larger $\sigma_1$ under MoCA indicates a stronger cross-view association: MoCA produces partitions that better exploit cross-modal correlations and thus support improved reconstruction, while the slice-wise constraint of synchronized masking suppresses informative cross-modal pairings within a time window. Because a larger $\sigma_1$ implies a smaller lower bound for $L_{\mathrm{MAE}}$, MoCA is preferable to synchronized masking when within-time slice correlations across modalities/axes are strong and should be leveraged, which is often the case in multi-modal wearable device data.

\end{document}